\documentclass[10pt,twocolumn]{IEEEtran} 	
\usepackage{fancyhdr, epsfig, epsf, amsthm, amsmath, amssymb, amsfonts, subfigure, color}
\usepackage{threeparttable}
\usepackage[noadjust]{cite}
\usepackage{dsfont}
\usepackage{enumerate}
\usepackage{comment}
\usepackage{soul}
\usepackage{algorithm}
\usepackage{algpseudocode}
\usepackage{mathtools}
\usepackage{textcomp}
\usepackage{gensymb}
\usepackage{array,booktabs,hhline,tabu}
\newcolumntype{P}[1]{>{\centering\arraybackslash}p{#1}}
\usepackage{amssymb}
\usepackage{multirow}
\usepackage{makecell}
\usepackage{adjustbox}
\usepackage{hyperref} 
\usepackage{cleveref}
\usepackage{diagbox}
\usepackage{hhline}
\Crefname{figure}{Fig.}{Figs.}
\usepackage{scalerel}
\usepackage{tikz}
\usepackage[free-standing-units=true]{siunitx}
\newcolumntype{C}{>{$}c<{$}}
\include{header}

\DeclarePairedDelimiterX{\norm}[1]{\lVert}{\rVert}{#1}

\theoremstyle{definition}
\newtheorem{definition}{Definition}
\theoremstyle{remark}
\newtheorem{remark}{Remark}
\theoremstyle{assumption}
\newtheorem{assumption}{Assumption}

\usepackage{scalerel}
\usepackage{tikz}
\usetikzlibrary{svg.path}

\definecolor{orcidlogocol}{HTML}{A6CE39}
\tikzset{
	orcidlogo/.pic={
		\fill[orcidlogocol] svg{M256,128c0,70.7-57.3,128-128,128C57.3,256,0,198.7,0,128C0,57.3,57.3,0,128,0C198.7,0,256,57.3,256,128z};
		\fill[white] svg{M86.3,186.2H70.9V79.1h15.4v48.4V186.2z}
		svg{M108.9,79.1h41.6c39.6,0,57,28.3,57,53.6c0,27.5-21.5,53.6-56.8,53.6h-41.8V79.1z M124.3,172.4h24.5c34.9,0,42.9-26.5,42.9-39.7c0-21.5-13.7-39.7-43.7-39.7h-23.7V172.4z}
		svg{M88.7,56.8c0,5.5-4.5,10.1-10.1,10.1c-5.6,0-10.1-4.6-10.1-10.1c0-5.6,4.5-10.1,10.1-10.1C84.2,46.7,88.7,51.3,88.7,56.8z};
	}
}

\newcommand\orcidicon[1]{\href{https://orcid.org/#1}{\mbox{\scalerel*{
				\begin{tikzpicture}[yscale=-1,transform shape]
				\pic{orcidlogo};
				\end{tikzpicture}
			}{|}}}}

\usepackage{hyperref} 
\hypersetup{ colorlinks=true, linkcolor=blue, filecolor=magenta, urlcolor=cyan }

\def\BibTeX{{\rm B\kern-.05em{\sc i\kern-.025em b}\kern-.08em
		T\kern-.1667em\lower.7ex\hbox{E}\kern-.125emX}}
\usepackage{balance}

\begin{document}
\title{Mobility-Induced Graph Learning for \\ WiFi Positioning}
\author{Kyuwon Han, Seung Min Yu, Seong-Lyun Kim, and Seung-Woo Ko 
\thanks{K. Han and S.-L. Kim are with Dept. EEE, Yonsei University, Korea (email: \{kwhan, slkim\}@ramo.yonsei.ac.kr). S. M. Yu is with Korea Railroad Research Institute, Korea (email: smyu@krri.re.kr). S.-W. Ko is with Dept. Smart Mobility Eng., Inha University, Korea (email: swko@inha.ac.kr).}}

\maketitle

\begin{abstract} A smartphone-based user mobility tracking could be effective in finding his/her location, while the unpredictable error therein due to low specification of built-in \emph{inertial measurement units} (IMUs) rejects its standalone usage but demands the integration to another positioning technique like WiFi positioning. This paper aims to propose a novel integration technique using a graph neural network called \emph{Mobility-INduced Graph LEarning} (MINGLE), which is designed based on two types of graphs made by capturing different user mobility features. Specifically, considering sequential \emph{measurement points} (MPs) as nodes, a user's regular mobility pattern allows us to connect neighbor MPs as edges, called \emph{time-driven mobility graph} (TMG). Second, a user's relatively straight transition at a constant pace when moving from one position to another can be captured by connecting the nodes on each path, called a \emph{direction-driven mobility graph} (DMG). Then, we can design \emph{graph convolution network} (GCN)-based cross-graph learning, where two different GCN models for TMG and DMG are jointly trained by feeding different input features created by WiFi RTTs yet sharing their weights. Besides, the loss function includes a mobility regularization term such that the differences between adjacent location estimates should be less variant due to the user's stable moving pace. Noting that the regularization term does not require ground-truth location, MINGLE can be designed under semi- and self-supervised learning frameworks. The proposed MINGLE's effectiveness is extensively verified through field experiments, showing a better positioning accuracy than benchmarks, say \emph{root mean square errors} (RMSEs) being $1.398$ (\si{\metre}) and $1.073$ (\si{\metre}) for self- and semi-supervised learning cases, respectively.       
\end{abstract}


\section{Introduction}\label{Sec:Intro}

With the widespread use of smartphones and ubiquitous WiFi \emph{access points} (APs), WiFi positioning has been considered a popular solution for providing user-specific services by identifying the user’s location \cite{yang2015wifi}. The distances between a target user and each WiFi AP can be estimated based on \emph{ signal strength} (SS) or \emph{round trip time} (RTT), allowing us to find the location estimate through multilateration. However, its accuracy and reliability are questionable since these measurements can be severely corrupted unless the \emph{line-of-sight} (LoS) condition is made between the user and each AP. One viable approach to overcome the limitation is that the user’s mobility information measured by a built-in \emph{inertial measurement unit} (IMU) is involved in WiFi positioning since it is independent of the propagation conditions affecting SS or RTT. Although various techniques to integrate user mobility into WiFi positioning have been proposed in the literature introduced in the sequel, this paper aims to propose a novel technique using a \emph{graph neural network} (GNN), called \emph{Mobility-INduced Graph LEarning} (MINGLE), which is based on graphs created by capturing user-mobility inherent features. Specifically, a user’s regular mobility patterns, such as constant moving speed and directions that are detectable by the user’s smartphone’s IMUs, enable us to generate graphs in real-time. Nodes represent the user’s locations, and the edges between the nodes are created depending on their temporal adjacency or direction consistency. The generated graphs are cores of the proposed MINGLE. The input features (raw or pre-processing data) are propagated to the following layers’ nodes only when the corresponding edges exist. Besides, a regular inter-distance between adjacent nodes can be used as a regularization term in designing a loss function, making it possible to develop MINGLE under semi-supervised and self-supervised learning frameworks.

\subsection{Prior Work}

We summarize relevant works in the literature, which can be divided into two parts: WiFi positioning with user mobility and GNN-based positioning. 

\subsubsection{WiFi Positioning with User Mobility}

\begin{table*}[!t]
\centering
\caption{Survey of GNN-based positioning methods}
\adjustbox{max width=\textwidth}{
\begin{tabular}{c|cccccc}
Reference                    & Node                                                       & Edge                                                                                       & Feature                                                               & GNN                    & Loss function                                                                          & Output               \\ \hline
\cite{sun2021novel}                 & APs                                                         & \begin{tabular}[c]{@{}c@{}}Fully-connected \\ (1/distance)\end{tabular}                    & SS                                                                    & GCN               & Cross-entropy                                                                          & Fingerprint label    \\ \hline
\cite{kang2023indoor}               & APs                                                         & \begin{tabular}[c]{@{}c@{}}Fully-connected \\ (1/distance)\end{tabular}                    & SS                                                                    & GCN               & MSE                     & 2D location          \\ \hline
\cite{luo2022geometric }            & APs, target                                                 & \begin{tabular}[c]{@{}c@{}}Fully-connected \\ (1/distance)\end{tabular}                    & SS                                                                    & GraphSAGE         & Huber                                                                                  & 2D location          \\ \hline
\multirow{3}{*}{\cite{wu2022multi}} & APs                                                         & \begin{tabular}[c]{@{}c@{}}Fully-connected \\ (1/distance)\end{tabular}                    & SS                                                                    & GIN  & \multirow{3}{*}{MAE} & 2D location          \\ \cline{2-5} \cline{7-7} 
                             & Clients                                                     & \begin{tabular}[c]{@{}c@{}}Fully-connected \\      (Distribution similiarity)\end{tabular} & \begin{tabular}[c]{@{}c@{}}Embedding \\ vector\end{tabular}           & GCN         &                                                                                        & Set of local weights \\ \hline
\cite{yan2021graph}                 & APs                                                         & \begin{tabular}[c]{@{}c@{}}Inter-distance\\      (Binary)\end{tabular}                     & \begin{tabular}[c]{@{}c@{}}Estimated\\ distance\end{tabular}          & GCN                      & RMSE                            & 2D location          \\ \hline
\cite{tang2023csi}                  & Pixels                                                      & \begin{tabular}[c]{@{}c@{}}Neighbor pixels\\      (Binary)\end{tabular}                    & \begin{tabular}[c]{@{}c@{}}Amplitude and \\ phase of CSI\end{tabular} & GraphSAGE        & Cross-entropy                                                                          & Fingerprint label    \\ \hline
MINGLE                       & \begin{tabular}[c]{@{}c@{}}Sequence \\ of MPs\end{tabular} & \begin{tabular}[c]{@{}c@{}}Temporal adjacency, \\ direction consistency\\ (Binary)\end{tabular}        & RTTs                                                                  &   GCN                    & MSE                                                                                    & 2D location         
\end{tabular}}
\label{Table:Survey}
\end{table*}

A recent smartphone equips various IMU sensors, e.g., an accelerometer,  gyroscope, and magnetometer, to help update the user's location in GPS-denied environments, called \emph{pedestrian dead reckoning} (PDR) \cite{kang2014smartpdr}. On the other hand, a standalone PDR approach is hardly considered due to several practical limitations, such as the low resolution of the built-in IMUs, accumulated error over time \cite{yun2023rampscope}, and time-varying holding patterns \cite{herath2020ronin}. It calls for integrating another positioning approach as a remedy to overcome the above limitations, e.g., WiFi positioning as a main target of this work.

The primary focus of research in this field is to create a method that combines WiFi positioning and PDR. One classic way is to use \emph{Kalman filter}  (KF) or \emph{extended KF} (EKF) where two different location estimates of WiFi positioning and PDR, termed \emph{prediction location estimate} (PLE) and \emph{measurement location estimate} (MLE) respectively, are combined by weighting a more reliable one \cite{sun2020indoor}. On the other hand, a prerequisite for KF- or EKF-based integration is that each location estimate's covariance matrix should be given in prior, yet challenging in practice. Several studies have been made to cope with this issue. 
In \cite{revach2022kalmannet} and \cite{choi2023split}, these covariance matrices are computed by leveraging the power of \emph{deep neural network} (DNN), assuming that the ground-truth covariance matrix is given for DNN training. In \cite{yu2021integrating}, a novel technique called \emph{combinatorial data augmentation} (CDA) is proposed for updating the MLE's covariance matrix without a ground-truth one such that spatial variance of multiple location estimates obtained from different subsets of WiFi RTT measurements is considered a real-time covariance matrix of MLE. Note that the above techniques are designed based on the belief that KF provides an optimal fusion between PLE and MLE, which only makes sense when the errors on PLE and MLE follow zero-mean Gaussian distributions. Besides, KF targets to fuse a pair of single PLE and MLE, making it difficult to distill useful information embedded in the previous results.

Unlike the above KF-based integration between a single PLE and MLE, a few recent studies aim to exploit the entire sequences of MLEs and PLEs. In \cite{han2021exploiting}, the above two sequences are aligned with the minimum error, called \emph{trajectory alignment} (TA), by jointly optimizing WiFi NLoS bias, step length, and initial heading direction. It is extended into the DNN-based algorithm in \cite{choi2022enhanced} that the loss function is defined as the difference between the two sequences. 
On the other hand, these two algorithms work well under the assumption of a relatively precise IMU measurement, necessitating precise calibration and pre-processing to this end \cite{han2023waveform}.

\subsubsection{GNN-Based Positioning}
GNN is a class of DNN that uses graph structures to extract domain knowledge therein \cite{xu2018powerful}. While conventional DNNs open all nodes to connect in the following layers, GNN compels each node to update its latent states based on the aggregated information from its adjacent nodes, effectively learning local features in large-scale networks. Traditionally, GNN has been applied to cases when a graph structure is innately given (e.g., social networking \cite{li2023survey}). On the other hand, several recent studies attempt to apply GNN to more general cases by creating a graph from the problem formulation, such as radio resource optimization in wireless networks \cite{shen2022graph}. 

Radio-based positioning is another emerging application that applies GNN, the central theme of this work. A straightforward way to create a graph is by designating positioning APs as nodes. Assuming that the APs' locations are given, the edges between them can be constructed based on their geometric relation. For example, a fully connected graph is created in \cite{sun2021novel} and \cite{kang2023indoor} whose edge weights are assigned inversely proportional to the corresponding inter-distances. The graph is used to construct a positioning algorithm comprising several \emph{graph convolution network} (GCN) layers, GNN version of convolution networks \cite{kipf2016semi}. 
In \cite{luo2022geometric}, three types of APs (WiFi, ZigBee, and Bluetooth) are considered, each modeled as a fully connected bidirectional graph. The directional edge between a user and an AP is then added, with a weight inversely proportional to the estimated distance. With the aid of GraphSAGE layer \cite{hamilton2017inductive}, it is possible to localize the user by unifying all generated graphs.
The scenarios with many users are considered in \cite{yan2021graph}, where both APs and users are designated nodes, assuming every node can communicate.  

Several recent studies have suggested different ways to configure a graph other than the above approach. 
In \cite{tang2023csi}, \emph{channel state information} (CSI) is converted into a 2D  image made up of complex numbers.  Pixels in the image are treated as nodes, and their adjacency determines connectivity. The complex number's magnitude and phase become an input feature of the constructed graph. In \cite{wu2022multi}, GNN-based localization is extended into federated learning, where the global server regards clients as nodes and determines an adjacency matrix between them by utilizing the similarity between different clients' embedding vectors. Each client updates its model based on \emph{graph isomorphism network} (GIN) \cite{xu2018powerful}, and the server aggregates them via GCN with the self-attention mechanism. We have summarized the GNN-based positioning techniques discussed above in Table \ref{Table:Survey}.

Despite their effectiveness, most GNN-based positioning techniques share one practical limitation: They are designed based on supervised learning. They require labeling many data samples, usually ground-truth locations at which the corresponding measurements are collected, which is a time-consuming and labor-intensive campaign. Besides, even if data labeling is given, 
the trained GNN's validity duration is limited because radio signals' features change over time due to various factors, e.g., temperature, pedestrian density, and so on, leading to a  mismatch between training data and actual measurements.

\subsection{Contributions}

\begin{figure}[t]
    \centering
    \includegraphics[width=9cm]{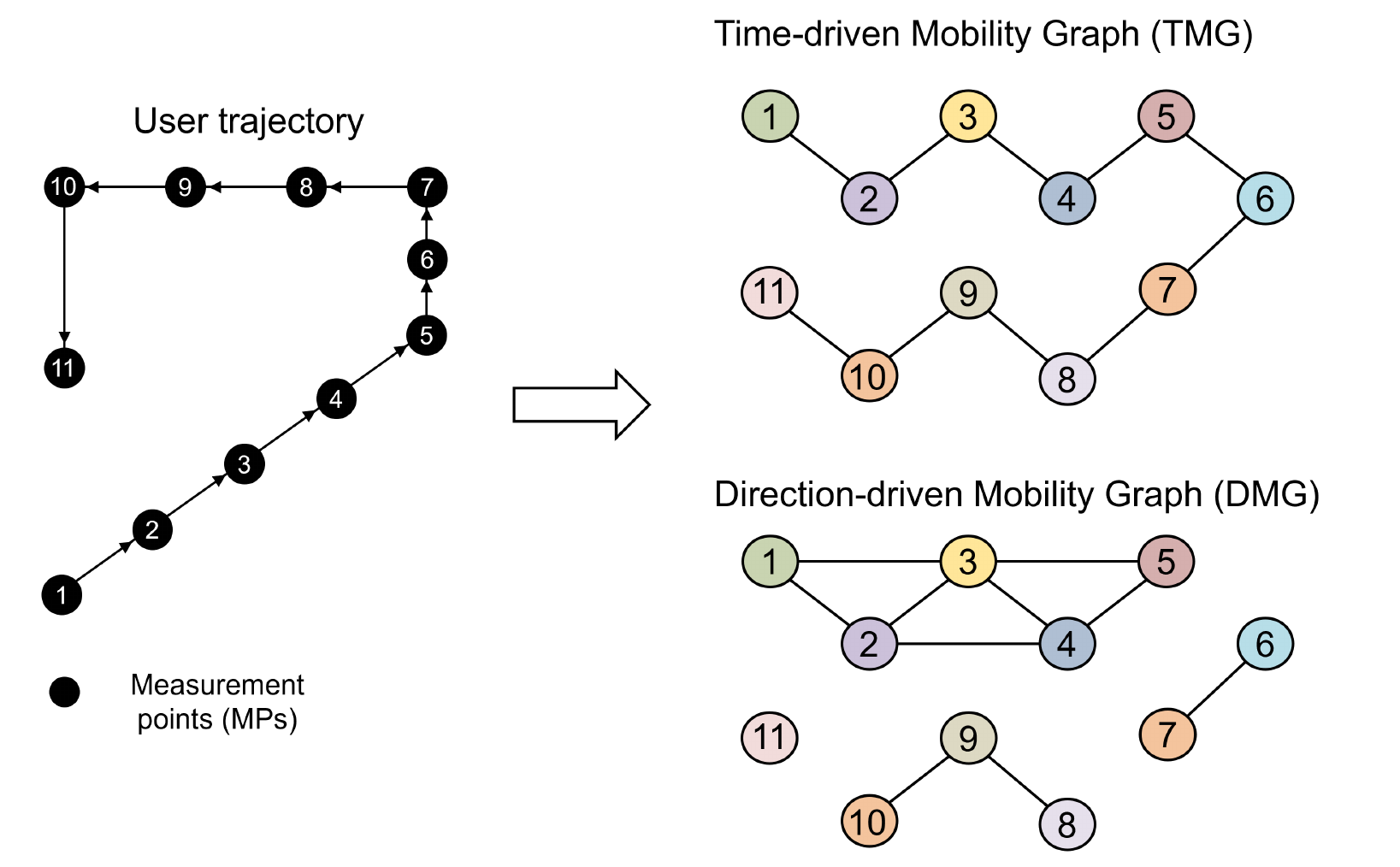}
    \caption{Graphical example representing the relation between a user's real trajectory and two types of mobility-induced graphs, namely,  TMG and DMG defined in Definition \ref{Def:MIG}.}
    \label{Fig:Mobility_graph}
\end{figure}

Contrary to the works mentioned above, we exploit user mobility as a critical ingredient to design MINGLE. Specifically, we focus on user mobility's inherent features: People tend to move at a relatively constant speed determined by their mobility types (e.g., walking, running, and so on) \cite{shim2016traffic} and make a  straight transition while moving from one to another place \cite{bettstetter2003node, rhee2011levy,soltani2020orientation}. Due to a smartphone's constant sampling interval, the displacement between adjacent \emph{measurement points} (MPs) usually remains invariant,  offering the following three-fold opportunities. First, considering the spatial correlation of WiFi measurements, we can create a mobility-induced graph by setting MPs as nodes and connecting them when they are temporally adjacent. We call it a \emph{time-driven mobility graph} (TMG) throughout the work. Next,  every MP on the user's straight transition, which is enough to be detectable using built-in IMUs, can be connected to create another mobility-induced graph, called a \emph{direction-driven mobility graph} (DMG). Figure \ref{Fig:Mobility_graph} shows a graphical illustration of TMG and DMG. As the user's trajectory becomes longer, the degrees of TMG and DMG increase, helping enhance the positioning performance. Last, such a regular displacement between neighbor MPs can be reflected in the loss-function design to train a GNN without the ground-truth location. In other words, exploiting user mobility makes it possible to develop MINGLE under the frameworks of semi-supervised learning and self-supervised learning, overcoming the main drawback of the state-of-the-art GNN positioning works.

The main contributions of this work are given below.

\begin{itemize}
    \item \textbf{Real-Time Generation of Mobility-Induced Graphs}: As mentioned before, two types of graphs, say TMG and DMG, are generated based on the real-time movement of the user, each effectively extracting distinct features that are useful when analyzing the user's mobility pattern. First, TMG is a connected graph that represents the inter-distance between MPs depending on the number of hops between the corresponding nodes. Second, DMG is a disconnected graph comprising multiple sub-graphs, each representing the user's straight-moving pattern. It helps express the directional consistency of MPs on each sub-graph. 
      \item \textbf{Cross-Graph Learning and Mobility Regularization of MINGLE Architecture}: MINGLE has two signature design components: cross-graph learning and mobility regularization. First, it is essential to extract different potentials of TMG and DMG and make a unified framework for estimating the user's location precisely. To this end, we design MINGLE based on cross-graph learning such that two different GNNs are individually trained with different types of input features while their weights are shared. Second, the user's stationary pace can be reflected on the loss function's regularization term, which is expressed as a small variance of inter-distances between adjacent MPs. As aforementioned, the mobility regularization does not require ground-truth locations and helps find precise location estimates of the user when a few or no ground-truths are given.

    \item \textbf{Extensive Field Experiment}: We conducted an extensive field experiment covering different mobility patterns and temporal environment changes. The proposed MINGLE's effectiveness is well verified by showing a much more accurate positioning result than several benchmarks, say the \emph{root mean square errors} (RMSEs) being $1.398$ and $1.073$ (\si{\metre}) without and with $5$-$10\%$ ground-truth locations, respectively.     
\end{itemize}

\section{System Model and Problem Formulation} \label{Sec:System_model}
This section explains our system model, including a  network model and types of measurements. Then, we describe the problem we tackle throughout the work.

\subsection{Network Model}\label{subsection:Network_Model}

Consider a WiFi positioning network in which $M$ WiFi APs are deployed. The set of APs' indices is denoted by $\mathcal{M}=\{1,2,\cdots, M\}$. Each AP's \emph{two-dimensional} (2D) coordinates, denoted by $\mathbf{z}^{(m)}\in\mathbb{R}^2$ for AP $m\in\mathcal{M}$, are assumed to be stationary and given in advance. Next, a smartphone user moves around the network, and several measurements explained in the sequel are sampled with the constant interval of $\Delta$ ($\sec$). The locations where the measurements are sampled are referred to as MPs. The sequence of the MPs' coordinates is given as $\{\mathbf{x}_n\}$, where $\mathbf{x}_n\in\mathbb{R}^2$ is the $n$-th MP's 2D coordinates. 
The set of MPs' indices is denoted by $\mathcal{N}=\{1,\cdots, N\}$, where $N$ is the number of MPs within the duration of $T$ ($\sec$), given as $N=\left\lfloor\frac{T}{\Delta}\right\rfloor$ with $\lfloor\cdot\rfloor$ being a floor function. Except for a few MPs that coincide with landmarks in the concerned area (e.g., pillars and gates), it is reasonable to consider the MP's coordinates to be generally unknown. We introduce a binary sequence $\{\alpha_n\}_{n=1}^{N}$ where $\alpha_n$ is one if the corresponding ground truth $\mathbf{x}_n$ is known while zero otherwise, say $\alpha_n\in\{0,1\}$. The index set of MPs without ground truths is defined as $\mathcal{X}=\{n|\alpha_n=0, n\in\mathcal{N}\}$.  
The objective of this work is to precisely estimate the coordinates of the MPs in $\mathcal{X}$.       

\subsection{Measurements and Assumption}

The user's smartphone can collect various measurements to help localize the user. Among them, this work considers two kinds of measurements, namely RTT and heading direction change, each of which will be elaborated as follows. 

\subsubsection{Round Trip Time}
An RTT between each WiFi AP and the smartphone, defined as double propagation time between the two, is measurable with the \emph{fine-timing measurement} (FTM) protocol initiated from IEEE 802.11mc \cite{IEEE2016}. We assume that both WiFi APs and the user's smartphone support the FTM protocol, and all APs' RTTs can be estimated at every sampled MP, denoted by $\boldsymbol\tau_n\in \mathbb{R}^{M}$ given as
\begin{align}\label{Eq:RTT}
\boldsymbol\tau_n=\left[\tau_n^{(1)},\tau_n^{(2)},\cdots, \tau_n^{(M)}\right], \quad n=1,\cdots, N,
\end{align}
where $\tau_n^{(m)}$ is the RTT measurement from WiFi AP $m$ at the $n$-th MP.   

With a LoS condition, the inter-distance between WiFi AP and the user is precisely detectable by multiplying the half of the light speed $c\approx 3\times 10^8$ (m/s) to the corresponding RTT, as experimentally verified in \cite{ibrahim2018verification}. On the other hand, in the case of an NLoS condition, a WiFi signal's propagation path is severely deviated from the direct path due to frequent blockages, resulting in  overestimating the inter-distance \cite{horn2020doubling}. Unlike several prior works addressing the NLoS issue by stochastic approaches (e.g., \cite{dong2021real} and  \cite{horn2020observation}), there are no statistics on LoS and NLoS conditions.

\begin{figure}[t]
    \centering
    \subfigure[User trajectory]{\includegraphics[height=3.6cm]{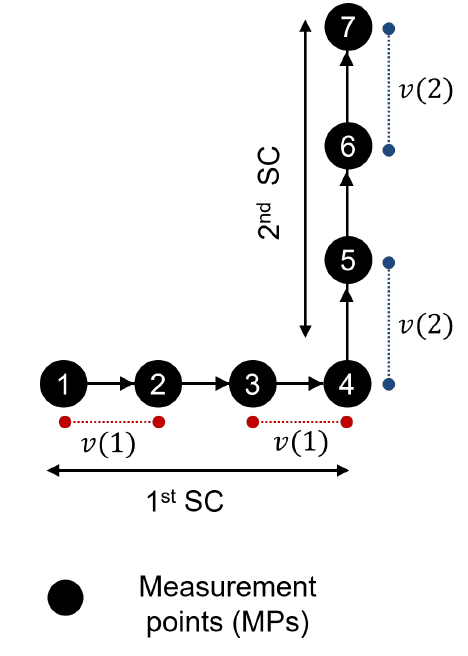}}
    \subfigure[IMU measurements]{\includegraphics[height=4.4cm]{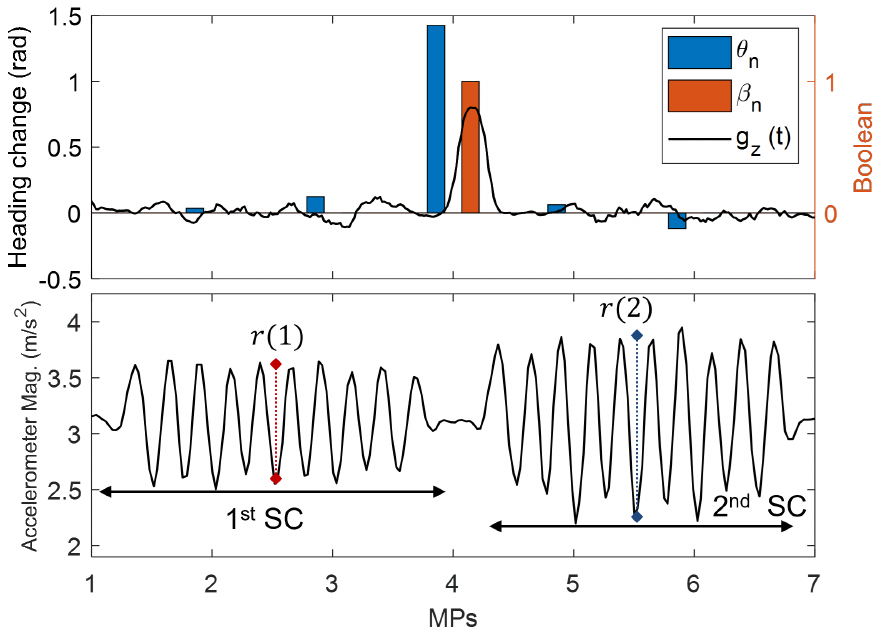}}
    \caption{The relation between user mobility trajectory and the resultant IMU measurements. (a) An example of user trajectory with $7$ MPs and $2$ SCs. (b) The accelerometer and gyroscope's measurements and their relevant metrics, including raw heading change $\theta_n$  \eqref{Eq:Theta_n}, binary heading change $\beta_n$  \eqref{Eq:Quantized_Heading_Change}, and the average gap between accelerometer' peaks and valleys $r(\ell)$ specified in \eqref{Eq:Speed_ratio}.}
    \label{Fig:Speed_variation}
\end{figure}

\subsubsection{Heading Direction Change} 
A smartphone's built-in gyroscope sensor can track its heading direction change. Consider an instant when the user is located at the $n$-th MP. During the movement to the $(n+1)$-th MP, the smartphone has recorded the gyroscope's z-axis measurement, denoted by $g_z(t)$ with $n\Delta < t \leq (n+1)\Delta$. 
Then, the corresponding heading change, denoted by $\theta_n$, can be computed by integrating 
$g_z(t)$~as
\begin{align} \label{Eq:Theta_n}
\theta_n=\int_{n\Delta}^{(n+1)\Delta} g_z(s) ds, \quad n=1,\cdots, N-1. 
\end{align}

On the other hand, the gyroscope's measurement is sensitive to the user's unpredictable activity, bringing about considerable bias and noise. The resultant heading change could be significantly different from the ground truth. Instead of  the noisy raw heading direction $\theta_n$, we quantize it into two states representing whether the direction is significantly changed. Specifically, we introduce a sequence of binary variable $\{\beta_n\}_{n=1}^{N}$, where  $\beta_n$ is the quantization value of $\theta_n$,   given as
\begin{align}\label{Eq:Quantized_Heading_Change}
\beta_n=\begin{cases} 1, \quad \text{if $\theta_n\geq \delta$}, \\
	0, \quad \text{otherwise,}
	\end{cases}
\end{align}
where $\delta$ is a predefined threshold. Without loss of generality, we set $\beta_N=1$ to represent the end of the trajectory. We set $\delta$ as $0.5$ (rad) in the field experiment in Sec. \ref{Sec:Verification}. 
Fig. \ref{Fig:Speed_variation} graphically illustrates the relation between raw heading change $\theta_n$ and its quantization value $\beta_n$. With $\{\beta_n\}$, we can make the following definition:
\begin{definition}[Steady Course]\label{Defintion: SC} A user is said to be on a \emph{steady course} (SC), which is defined as the sub-sequence of consecutive MPs whose $\beta_n$ therein are all zero except the last one. Specifically, the $\ell$-th SC can be expressed in terms of the sub-sequence $\{s(\ell),s(\ell)+1,\cdots, e(\ell)-1, e(\ell)\}$, satisfying the following condition:
\begin{align}\label{eq:SC}
\{\beta_{s(\ell)},\beta_{s(\ell)+1},\cdots, \beta_{e(\ell)-1}, \beta_{e(\ell)} \}=\{0,0,\cdots,0,1\},
\end{align}
where $s(\ell)$ and $e(\ell)$ are the first and the last MPs' indices on the $\ell$-th SC, respectively.
\end{definition}

\begin{assumption}[Consistent Speed]\label{Assumption: Consistent_Speed} \emph{The user's moving speed is assumed to be constant on the same SC, while it may be changed over different SCs.\footnote{As mentioned in Sec. \ref{Sec:Intro}, a user tends to move from one to another place in a consistent moving pattern (walking, running, and so on) with a stationary pace, allowing many works in the literature to use Assumption \ref{Assumption: Consistent_Speed} (see, e.g., \cite{bettstetter2003node},\cite{rhee2011levy}, and \cite{soltani2020orientation}).}} 
\end{assumption}
Assumption \ref{Assumption: Consistent_Speed} allows us to consider the  MPs on the same SC equally separated. On the other hand, the relation of MPs' inter-distance on different SCs will be explained in the following part.

\subsubsection{Speed Variation}\label{Sec:SpeedVariation}
A smartphone's accelerometer helps estimate how much its user accelerates or decelerates his/her pace when changing SCs. Denote the accelerometer's three-dimensional measurement $\boldsymbol{u}(t)\in \mathbb{R}^{3}$, whose $\mathsf{L2}$-norm $\|\boldsymbol{u}(t)\|$ repeatedly oscillates between peaks and valleys when moving on the $\ell$-th SC, as shown in Fig. \ref{Fig:Speed_variation}. By following several prior studies, e.g., \cite{kang2014smartpdr} and \cite{wu2018pedestrian}, the average gap between peaks and valleys within the same SC, denoted by $r(\ell)$, is considered inversely proportional to the fourth power of the step length. Noting that the step length is proportional to the corresponding moving speed, the moving speed on the $\ell$-th SC can be given as $\eta \cdot r(\ell)^{\frac{1}{4}}$
where $\eta$ depends on the user's personal parameter unknown in practice. Then, we can obtain the speed variation between different SCs by normalizing it with  their minimum value, given as
\begin{align}\label{Eq:Speed_ratio}
v(\ell)=\frac{r(\ell)^{\frac{1}{4}}}{\min_s\left[ {r(s)}^{\frac{1}{4}}\right]},
\end{align}
which is independent of $\eta$.

\begin{figure*}[t]
    \centering
    \includegraphics[width=15cm]{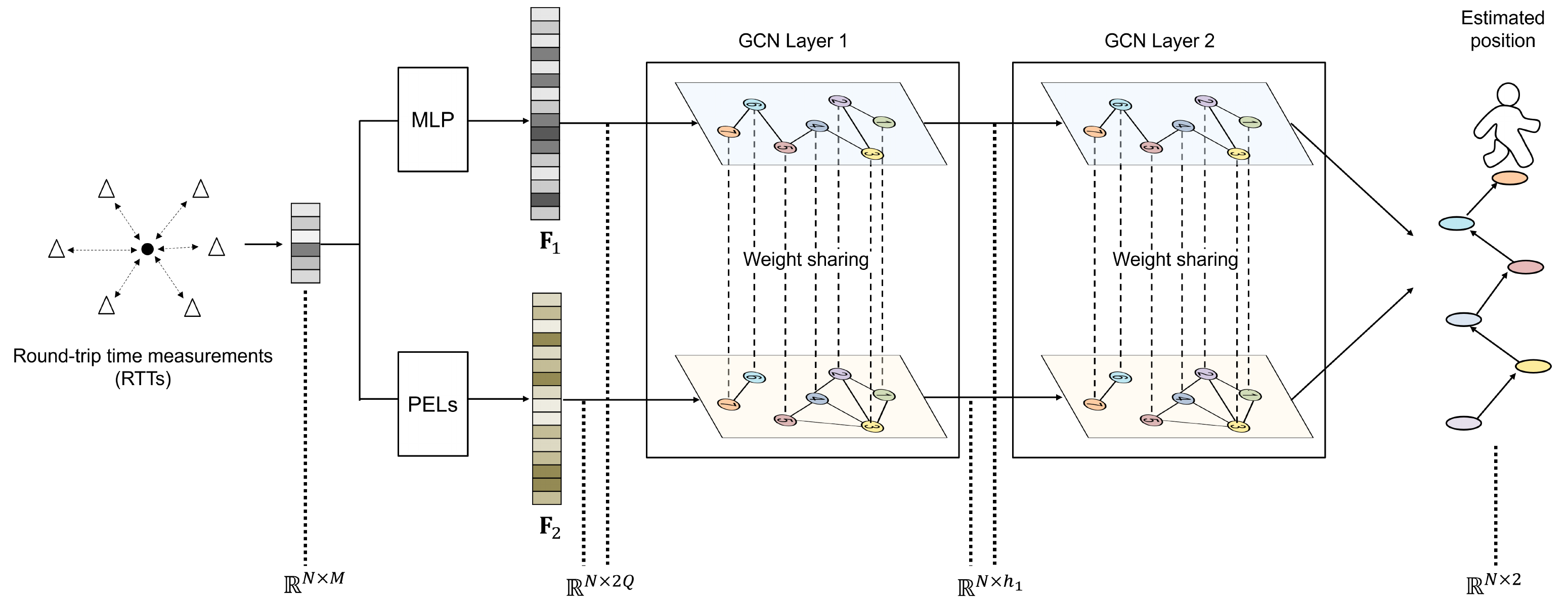}
    \caption{Graphical illustration of MINGLE architecture.}
    \label{Fig:Mingle}
\end{figure*}

\subsection{Problem Description}\label{sec:Problem_Def}

Given $\{\boldsymbol\tau_n\}$ of \eqref{Eq:RTT}, $\{(s(\ell), e(\ell))\}$ of \eqref{eq:SC}, and $\{v(\ell)\}$ of \eqref{Eq:Speed_ratio}, we attempt to estimate the user's location in semi-supervised or self-supervised learning manner. Specifically, the sequence of the location estimates  is denoted by $\{\hat{\mathbf{x}}_n\}$, where $\hat{\mathbf{x}}_n$ represents the location estimate at the $n$-th MP. As recalled in Sec. \ref{subsection:Network_Model}, the coordinates of all APs, say $\mathbf{z}^{(m)}$ for all $m\in \mathcal{M}$, are known in advance. Conditioned on the above information, we target to find a mapping function $f$ defined as
\begin{align}
\{\hat{\mathbf{x}}_n\}=f\left(\{\boldsymbol\tau_n\}, \{(s(\ell),e(\ell))\}, \{v(\ell)\} \left|\{\mathbf{z}^{(m)}\}\right.\right),
\end{align}
which is constrained by the following three requirements:
\subsubsection{Ground-Truth Location} The location estimates at the MPs not in $\mathcal{X}$ should be equivalent to the ground truth, namely, 
\begin{align} \label{Eq:Constraint1}
\hat{\mathbf{x}}_n=\mathbf{x}_n, \quad \forall n\in \mathcal{X}^c. 
\end{align}
Unless the complementary set $\mathcal{X}^c$ is empty, the concerned positioning problem becomes semi-supervised learning. Otherwise, it is reduced to self-supervised learning.   
\subsubsection{Equivalent Inter-Distance within one SC} By Assumption \ref{Assumption: Consistent_Speed}, the location estimates of adjacent MPs on the same SC should be equally separated, namely,  
\begin{align} \label{Eq:Constraint2}
\|\hat{\mathbf{x}}_{n-1}-\hat{\mathbf{x}}_{n}\|=\|\hat{\mathbf{x}}_{n}-\hat{\mathbf{x}}_{n+1}\|,\quad s(\ell)\leq n < e(\ell), \quad n\geq 2,
\end{align}
where $s(\ell)$ and $e(\ell)$ are the first and last MPs' indices on the $\ell$-th SC specified in \eqref{eq:SC}. 
\subsubsection{Inter-Distance Change over SCs}
As recalled in Sec. \ref{Sec:SpeedVariation}, the inter-distance between adjacent MPs on the $\ell$-th SC is proportional to $v(\ell)$ of \eqref{Eq:Speed_ratio}, leading to the following condition:
\begin{align}\label{Eq:Constraint3}
\frac{\|\hat{\mathbf{x}}_{n_1-1}-\hat{\mathbf{x}}_{n_1}\|}{v(\ell_1)}=\frac{\|\hat{\mathbf{x}}_{n_2-1}-\hat{\mathbf{x}}_{n_2}\|}{v(\ell_2)},
\end{align}
where $s(\ell_1)\leq n_1\leq e(\ell_1)$ and $s(\ell_2)\leq n_2\leq e(\ell_2)$ for $n_1\geq 2$ and $n_2\geq 2$.

\section{Mobility-Induced Graph Learning} \label{Sec:Proposed method}

This section proposes a novel GNN-based positioning algorithm, called MINGLE, to address the problem mentioned in Sec. \ref{sec:Problem_Def}. We first introduce two types of graphs induced by user mobility and overview the entire architecture of MINGLE based on the generated graphs. Next, we will explain each part of MINGLE in detail, including input feature generation, network architecture, and loss function. 

\begin{figure*}[t]
    \centering
    \includegraphics[width=15cm]{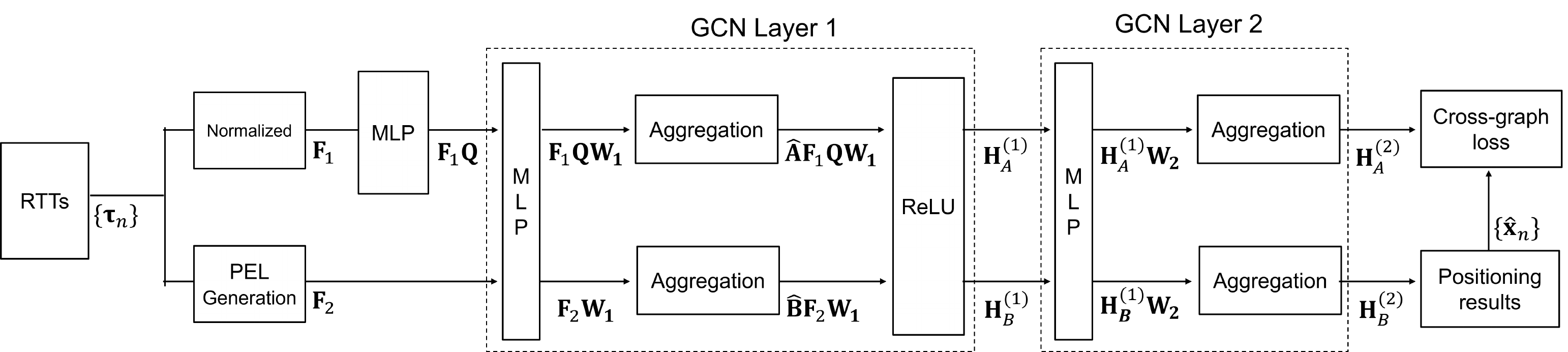}
    \caption{Two-layer GNN architecture of MINGLE, each of which the layer comprises two GCNs for TMG and DMG with sharing their weights.}
    \label{Fig:Network_architercture}
\end{figure*}

\subsection{Mobility-Induced Graphs and Architecture Overview} \label{Subsec:Graph_modeling}
A graph can be defined when nodes and edges are specified. In the following definition, two generated graphs, say TMG and DMG, will be defined from the perspective of their nodes and edges. 
\begin{definition} [Mobility-Induced Graphs]\label{Def:MIG} TMG and DMG share the same nodes as all $N$ MPs' indices $\mathcal{N}$. On the other hand, the edges of TMG and DMG  depend on adjacent matrices $\mathbf{A}, \mathbf{B} \in\mathbb{R}^{N\times N}$, respectively, as specified below:
\begin{itemize}
    \item \textbf{TMG}: Two nodes in TMG are linked if the difference between the two indices is smaller than the predefined threshold $\epsilon$, given as
\begin{align} \label{Eq:Mobility_adjacency}
	\mathbf{A}_{i,j}= \begin{cases} 1, \quad \text{if}\,\, |i - j| \leq \epsilon, \\
	0, \quad \text{otherwise.}
	\end{cases}
\end{align}
\item \textbf{DMG}: The edges of DMG depend on the SC defined in Definition~\ref{Defintion: SC}. Assume there are $L$ SCs, each of which the length is given as {$d(\ell)=e(\ell)-s(\ell)+1$}. Then, 
two nodes of DMG are linked if their difference is less than the half-length, given as $\mathbf{B}=\sum_{\ell=1}^L\mathbf{B}(\ell)$, where

\begin{align}
    \left(\mathbf{B}(\ell)\right)_{i,j} =\begin{cases} 1,  & \text{if $i,j\in\{s(\ell),\cdots,e(\ell)\}$} \\  & \quad \quad \text{and} \, |i - j| \leq \lfloor\frac{d(\ell)}{2}\rfloor,\\
	0, & \text{otherwise}. 
	\end{cases}
\end{align}

\end{itemize}
As a result, TMG and DMG can be represented as $\mathsf{G}_A(\mathcal{N},\mathcal{E}_A)$ and $\mathsf{G}_B(\mathcal{N},\mathcal{E}_B)$ respectively, where
\begin{align}
\mathcal{E}_A=\{(i,j)|\mathbf{A}_{i,j}=1\}, \quad
\mathcal{E}_B=\{(i,j)|\mathbf{B}_{i,j}=1\}.
\end{align}
\end{definition}
\begin{remark} [TMG and DMG]\label{remark:TMG_DMG}
The two generated graphs can contribute differently to estimating the user's unknown locations. TMG effectively captures the spatial correlation between neighbor MPs sampled at similar times, helping every location estimate connected to its neigbor ones. On the other hand, DMG focuses on the user's linear movements, relieving the burden of localizing the MPs thereon by reducing the concerned searching dimensions.
\end{remark}

To fully unleash the distinct potentials of TMG and DMG, we develop the architecture of MINGLE based on them, as illustrated in Fig. \ref{Fig:Mingle}. Each part of the architecture will be elaborated in the following subsections.

\subsection{Input Feature Generation}

We use two types of input features. The first is the matrix stacking normalized RTT vectors, denoted by $\mathbf{F}_1\in\mathbb{R}^{N\times M}$, given as
\begin{align}\label{Eq:F1}
\mathbf{F}_1=\left[\frac{\boldsymbol\tau_1}{\boldsymbol\tau_1 \mathbf{1}};\frac{\boldsymbol\tau_2}{\boldsymbol\tau_2 \mathbf{1}} ;\cdots ;\frac{\boldsymbol\tau_N}{\boldsymbol\tau_N \mathbf{1}} \right],
\end{align}
where $\mathbf{1}$ is a column vector whose elements are all $1$. 

Second, motivated by the CDA approach in \cite{yu2021integrating}, we augment the above input feature through different RTT combinations. For example, consider  $M$ RTTs at the $n$-th MP, say $\boldsymbol\tau_n$ of \eqref{Eq:RTT}. We can choose $K$  among $M$ ones, enabling us to make $Q=\binom{M}{K}$ different RTT sets. When $K\geq 3$,  each set's RTTs can be transformed into a 2D position estimate using a conventional multilateration algorithm (e.g., \cite{LLS_RS2008}), denoted by $\mathbf{y}^{(q)}\in \mathbb{R}^{1\times 2}$. We call it a \emph{preliminary estimated location} (PEL). We make the augmented input feature vector $\boldsymbol{f}_n\in\mathbb{R}^{1\times 2Q}$ by concatenating all PELs, given as
\begin{align} \label{Eq:f_n}
\boldsymbol{f}_n=\left[\mathbf{y}^{(1)},\mathbf{y}^{(2)},\cdots, \mathbf{y}^{(Q)}\right]. 
\end{align}
Then, the second input feature matrix, denoted by $\mathbf{F}_2\in\mathbb{R}^{N\times 2Q}$ can be derived by stacking the normalized version of  $\boldsymbol{f}_n$, given as
\begin{align}
\mathbf{F}_2=\left[\frac{\boldsymbol{f}_1}{\boldsymbol{f}_1 \boldsymbol{1} };\frac{\boldsymbol f_2}{\boldsymbol {f}_2\boldsymbol{1}} ; \cdots ; \frac{\boldsymbol f_N}{\boldsymbol {f}_N\boldsymbol{1}}\right].
\end{align}

\begin{remark}[Low-Dimensional vs. High-Dimensional Features]\label{remark:Input_Feature} Noting that the number of PELs $Q$ is much larger than the number of APs $M$, two matrices $\mathbf{F}_1$ and $\mathbf{F}_2$ can be considered the input features embedded in low-dimensional and high-dimensional spaces, respectively, which play complementary roles in the proposed GNN-based positioning algorithm. GNN is known to be effective when large-scale input features are given, like $\mathbf{F}_2$. On the other hand, the augmentation to a high dimensional space may diminish the spatial correlation between neighbor features. We use $\mathbf{F}_1$ as a supplementary to address the limitation, which is low-dimensional but preserves the spatial locality of the RTT measurements. 
\end{remark}

\subsection{Network Architecture: Cross-Graph Design}\label{Sec:Cross_Graph}

This subsection explains our network architecture based on cross-graph learning, where two different GCN models for TMG $\mathsf{G}_A(\mathcal{N},\mathcal{E}_A)$ and DMG $\mathsf{G}_B(\mathcal{N},\mathcal{E}_B)$ are jointly trained by sharing their weights. To this end, we use a two-layer GCN model for the two graphs, as illustrated in Fig. \ref{Fig:Network_architercture}. Specifically, the first layer's  vector representations, whose dimensions are set to $h_1$, are a function of a weight matrix $\mathbf{W}_1\in\mathbb{R}^{2Q\times h_1}$, given as
\begin{align}
\mathbf{H}_A^{(1)}\left(\mathbf{W}_{1}\right| \mathsf{G}_A) &\,= \mathsf{ReLU}(\hat{\mathbf{A}}  \mathbf{F}_{1}  \mathbf{Q}  \mathbf{W}_{1}), \\
\mathbf{H}_B^{(1)}\left(\mathbf{W}_{1}\right| \mathsf{G}_B) &\,= \mathsf{ReLU}(\hat{\mathbf{B}}  \mathbf{F}_{2}  \mathbf{W}_{1}).
\end{align}
Here, $\hat{\mathbf{A}}\in\mathbb{R}^{N\times N}$ and $\hat{\mathbf{B}}\in\mathbb{R}^{N\times N}$ are the normalized adjacency matrices, given as
\begin{align}
\hat{\mathbf{A}}=\mathbf{D}_A^{-\frac{1}{2}}\mathbf{A}\mathbf{D}_A^{-\frac{1}{2}}, \quad \hat{\mathbf{B}}=\mathbf{D}_B^{-\frac{1}{2}}\mathbf{B}\mathbf{D}_B^{-\frac{1}{2}}, 
\end{align}
where $\mathbf{D}_A$ and $\mathbf{D}_B$ are diagonal matrices whose elements represent the sum of the corresponding row vectors in $\mathbf{A}$ and $\mathbf{B}$, respectively. One observes that the input features $\mathbf{F}_1$ and $\mathbf{F}_2$ are separately fed into the GCN models for TMG and DMG, respectively. This mapping between the generated graphs and input features aligns well with the discussions in Remarks \ref{remark:TMG_DMG} and \ref{remark:Input_Feature}, the optimality of which will be experimentally studied in Sec. \ref{Sec:Verification}. Besides, we multiply \emph{multi-layer perceptron} (MLP) weight matrix $\mathbf{Q}\in \mathbb{R}^{M\times 2Q}$ with $\mathbf{F}_1$. The resultant dimension then becomes equivalent to $\mathbf{F}_2$, facilitating the sharing of the common weight~$\mathbf{W}_1$.

Next, the above first layer's outputs are recursively inputted to the second GCN layers, whose common weight matrix is denoted by $\mathbf{W}_{2}\in\mathbb{R}^{h_1\times 2}$. Then, the resultant embedding vectors,
say $\mathbf{H}_A^{(2)}\in\mathbb{R}^{N\times 2}$ and $\mathbf{H}_B^{(2)}\in\mathbb{R}^{N\times 2}$, are functions of both $\mathbf{W}_{1}$ and $\mathbf{W}_{2}$,
given as
\begin{align} \label{Eq:H_A}
\mathbf{H}_A^{(2)}\left(\mathbf{W}_{1}, \mathbf{W}_{2}\right| \mathsf{G}_A) &\,= \hat{\mathbf{A}}\mathbf{H}_A^{(1)}    \mathbf{W}_{2}\nonumber\\
&=\hat{\mathbf{A}}\left(\mathsf{ReLU}(\hat{\mathbf{A}}  \mathbf{F}_{1}  \mathbf{Q}  \mathbf{W}_{1})\right)    \mathbf{W}_{2},\\ \label{Eq:H_B}
\mathbf{H}_B^{(2)}\left(\mathbf{W}_{1}, \mathbf{W}_{2}\right| \mathsf{G}_B) &\,= \hat{\mathbf{B}}  \mathbf{H}_{B}^{(1)}  \mathbf{W}_{2}\nonumber\\
&=\hat{\mathbf{B}}  \left(\mathsf{ReLU}(\hat{\mathbf{B}}  \mathbf{F}_{2}  \mathbf{W}_{1})\right)  \mathbf{W}_{2}.
\end{align}
As a result, the RTT measurements at the $n$-th MP, say $\boldsymbol{\tau}_n$ of \eqref{Eq:RTT}, can be transformed to two vector representations embedded into 2D space, given as    
\begin{align} \label{Eq:Vector_representations}
\mathbf{a}_n=\mathsf{r}_n(\mathbf{H}_A^{(2)}), \quad  \mathbf{b}_n=\mathsf{r}_n(\mathbf{H}_B^{(2)}), 
\end{align}
where $\mathsf{r}_n(\cdot)$ returns the $n$-th row vector in the matrix therein. We aim to train the weight matrices $\mathbf{W}_1$ and $\mathbf{W}_2$ to generate the final vector representations $\{\mathbf{a}_n\}$ and $\{\mathbf{b}_n\}$ that should be similar to the ground-truth locations $\{\mathbf{x}_n\}$ by minimizing the loss function introduced in the following subsection.

\subsection{Loss Function Design}
By concerning three conditions specified in \eqref{Eq:Constraint1}, \eqref{Eq:Constraint2}, and \eqref{Eq:Constraint3}, we design a loss function, denoted by $\mathsf{Loss}(\{\mathbf{a}_n\}, \{\mathbf{b}_n\})$, which comprises the two terms representing the RMSE of the location estimates and their inter-distances regularized by the user mobility.

\subsubsection{Mean Square Error} The MSE of the location estimates can be divided into two cases with and without ground-truth locations.
First, consider a few MPs whose  ground truth locations are given, say $\mathbf{x}_n$ with $n\in \mathcal{X}^c$. By inputting the vector representations of \eqref{Eq:Vector_representations}, we can define the following MSE-based loss function:  
\begin{align} \label{Eq:L1} 
	\mathsf{MSE}_1(\{(\mathbf{a}_n,\mathbf{b}_n)\}_{n\in\mathcal{X}^c})=   \sum_{n\in\mathcal{X}^c} \norm{\mathbf{a}_n - \mathbf{x}_n}^2+\norm{\mathbf{b}_n - \mathbf{x}_n}^2,
\end{align}
whose minimization leads to satisfying the first condition of~\eqref{Eq:Constraint1}.

Second, inspired by \cite{yu2021integrating}, we can use the CDA-based location estimates as their real-time labels. Despite the errors on such labels, it has been experimentally verified that decent positioning accuracy is achievable, even superior to the labels themselves. Denote $\mathbf{c}_n$ a CDA-based location estimate at the $n$-th MP. For brevity, the detailed algorithm to derive $\mathbf{c}_n$ is omitted but summarized in Appendix \ref{Appendix:CDA}. Given $\{\mathbf{c}_n\}$, we can compute an MSE-based loss function for the MPs without ground truths, given as  
\begin{align} \label{Eq:L2} 
	\mathsf{MSE}_2(\{(\mathbf{a}_n,\mathbf{b}_n)\}_{n\in\mathcal{X}})=   \sum_{n\in\mathcal{X}} \norm{\mathbf{a}_n - \mathbf{c}_n}^2+\norm{\mathbf{b}_n - \mathbf{c}_n}^2.
\end{align}
By equally weighting \eqref{Eq:L1} and \eqref{Eq:L2}, we can obtain the overall MSE-based loss function, given as
\begin{align}\label{Loss_MSE}
\mathsf{Loss}_{\text{MSE}}(\{\mathbf{a}_n,\mathbf{b}_n\})=&\frac{1}{2}\mathsf{MSE}_1(\{(\mathbf{a}_n,\mathbf{b}_n)\}_{n\in\mathcal{X}^c})\nonumber\\
&+\frac{1}{2} \mathsf{MSE}_2(\{(\mathbf{a}_n,\mathbf{b}_n)\}_{n\in\mathcal{X}}).
\end{align}

\subsubsection{Mobility Regularization}
Recalling the assumption of the constant speed on the same SC in Assumption \ref{Assumption: Consistent_Speed}, we formulate the problem of minimizing the variance of the inter-distance between adjacent MPs, which is a surrogate for the constraints \eqref{Eq:Constraint2} and \eqref{Eq:Constraint3}. 
Given the vector representation $\mathbf{b}_n$ in \eqref{Eq:Vector_representations}, we denote $\gamma_n$ the inter-distance between the $(n-1)$-th and $n$-th MPs' location estimates normalized by the speed variation factor $v(\ell)$ of \eqref{Eq:Speed_ratio} for $ n \geq 2$, namely,
\begin{align}
\gamma_{n}=
\frac{\norm{\mathbf{b}_{n} - \mathbf{b}_{n-1}}}{v(\ell^*)},\quad \ell^*=\{\ell| n\in\{s(\ell),\cdots, e(\ell)\}\}. 
\end{align}

Then, the corresponding loss function is given as
\begin{align} \label{Eq:Mobility_regularization} 
        \mathsf{Loss}_{\text{MR}}(\{\mathbf{b}_n\})= \mathsf{var} (\{\gamma_n\}),
\end{align}
where $\mathsf{var}(\cdot)$ represents the variance of the sequence therein. 

By summing up the two loss functions \eqref{Loss_MSE} and \eqref{Eq:Mobility_regularization}, we can derive the overall loss function as
\begin{align}\label{Eq:Loss_function} 
\mathsf{Loss}(\{\mathbf{a}_n\}, \{\mathbf{b}_n\})=&\frac{1}{1+\lambda} \mathsf{Loss}_{\text{MSE}}(\{\mathbf{a}_n,\mathbf{b}_n\}) \nonumber \\ &+ \frac{\lambda}{1+\lambda} \mathsf{Loss}_{\text{MR}}(\{\mathbf{b}_n\}),
\end{align}
where $\lambda\geq0$ is a weight parameter. After the convergence of the loss function \eqref{Eq:Loss_function}, we use the second vector representation $\{\mathbf{b}_n\}$ in \eqref{Eq:Vector_representations} as the location estimates, namely,
\begin{align} \label{Eq:Final_estimation}
\hat{\mathbf{x}}_n=\mathbf{b}_n, \quad \forall n\in\mathcal{N},
\end{align}
whose optimality has been extensively studied through field experiments and shown to be more accurate than the other vector representation $\{\mathbf{a}_n\}$ in Sec. \ref{Sec:Cross-graph_validation}.  

\begin{figure*}[t]
	\centering
	\subfigure [Snapshots of the experiments.]{ 
        \subfigure{\includegraphics[height=4cm]{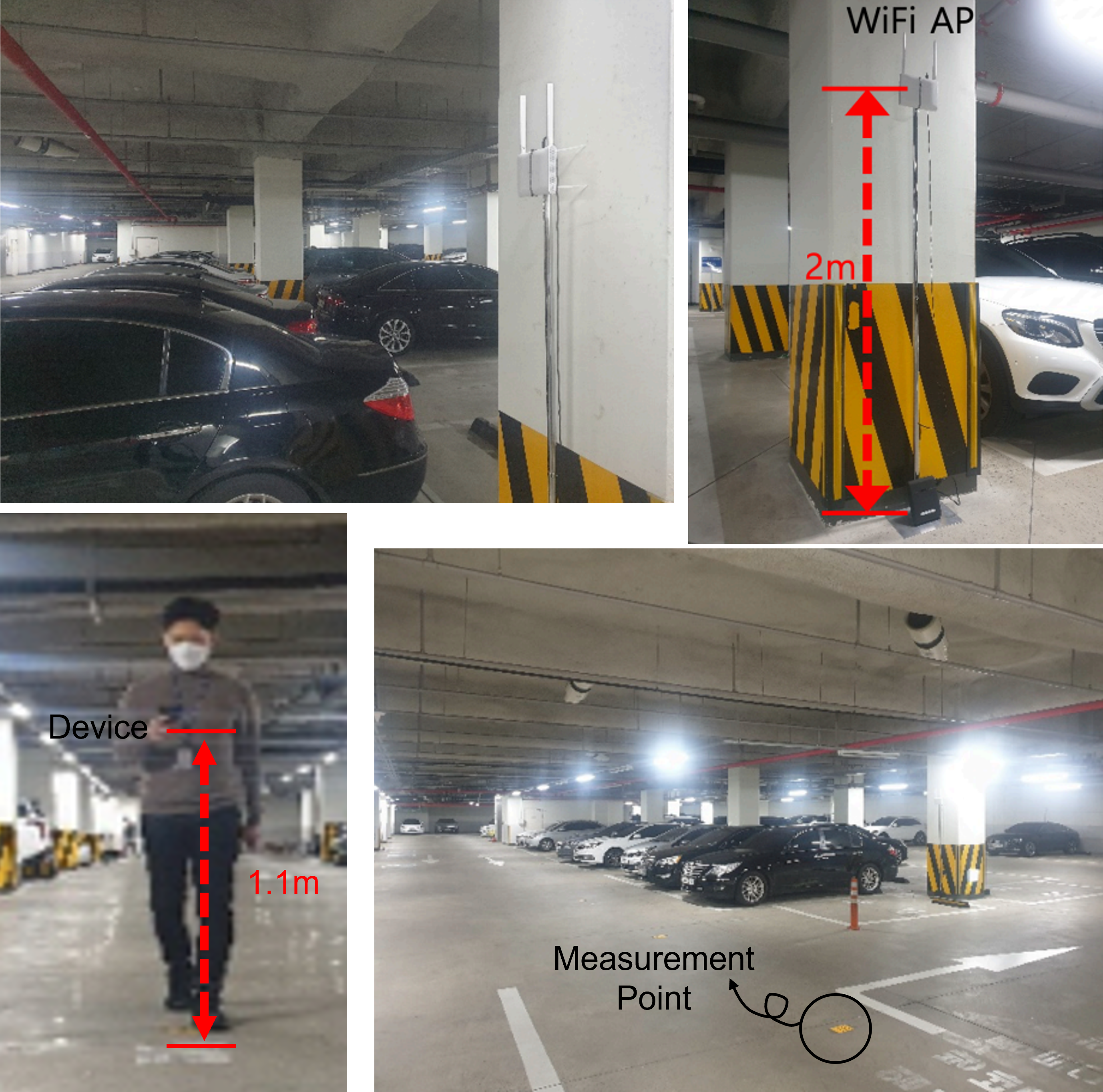}}
        \setcounter{subfigure}{1}
        }
        \subfigure[Experiment environment scanned by a LIDAR.]{ 
        \subfigure{\includegraphics[height=4cm]{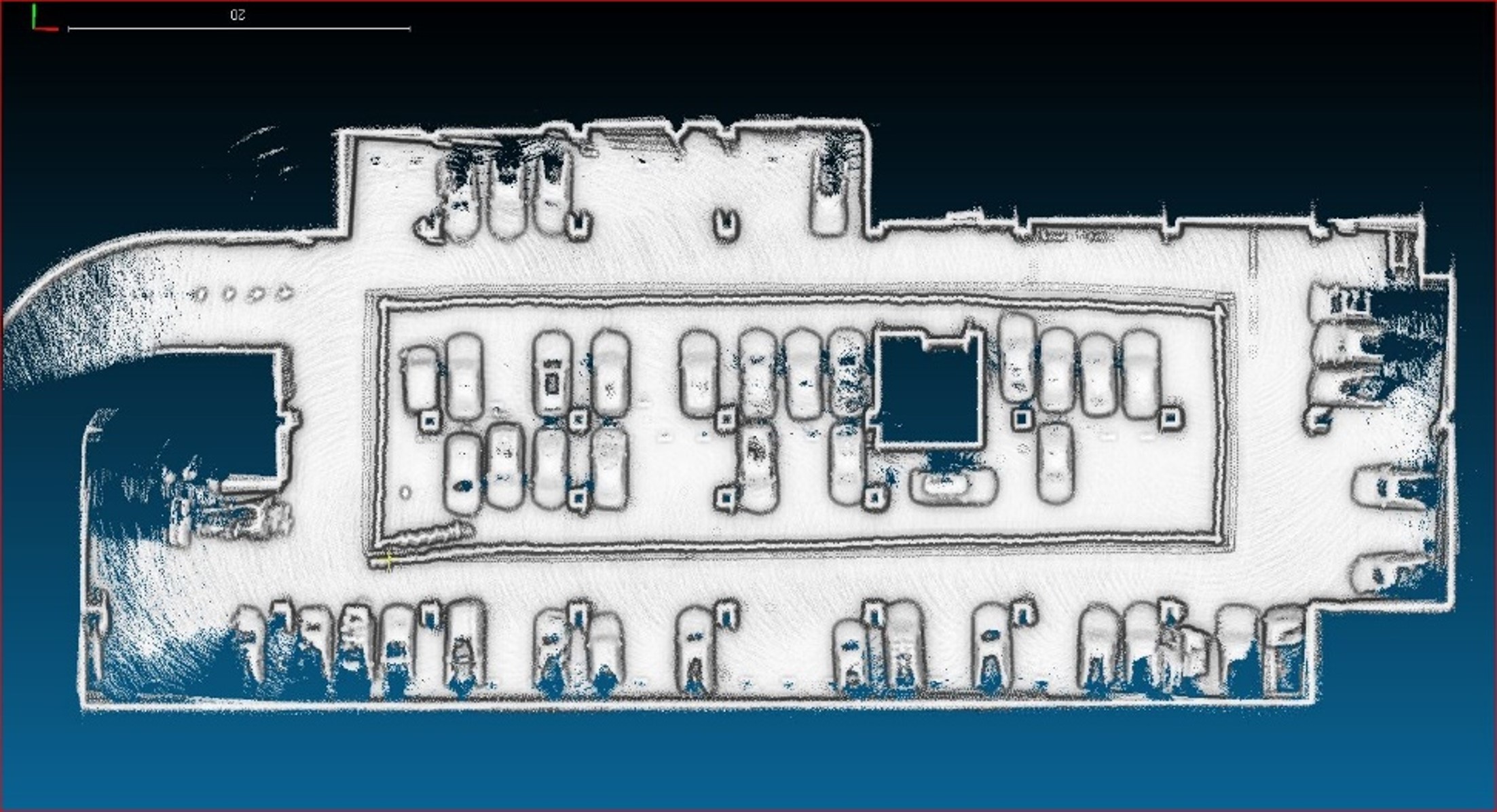}} 
        \setcounter{subfigure}{2}
        }
	\caption{Several photos of our experiment site and its floor plan scanned by a LIDAR. Parked cars' locations vary from day to day.}
	\label{Fig:Test_environments}
\end{figure*}

\begin{figure*}[t]
	\centering
	\subfigure[Type $1$.]{\includegraphics[width=8.0cm]{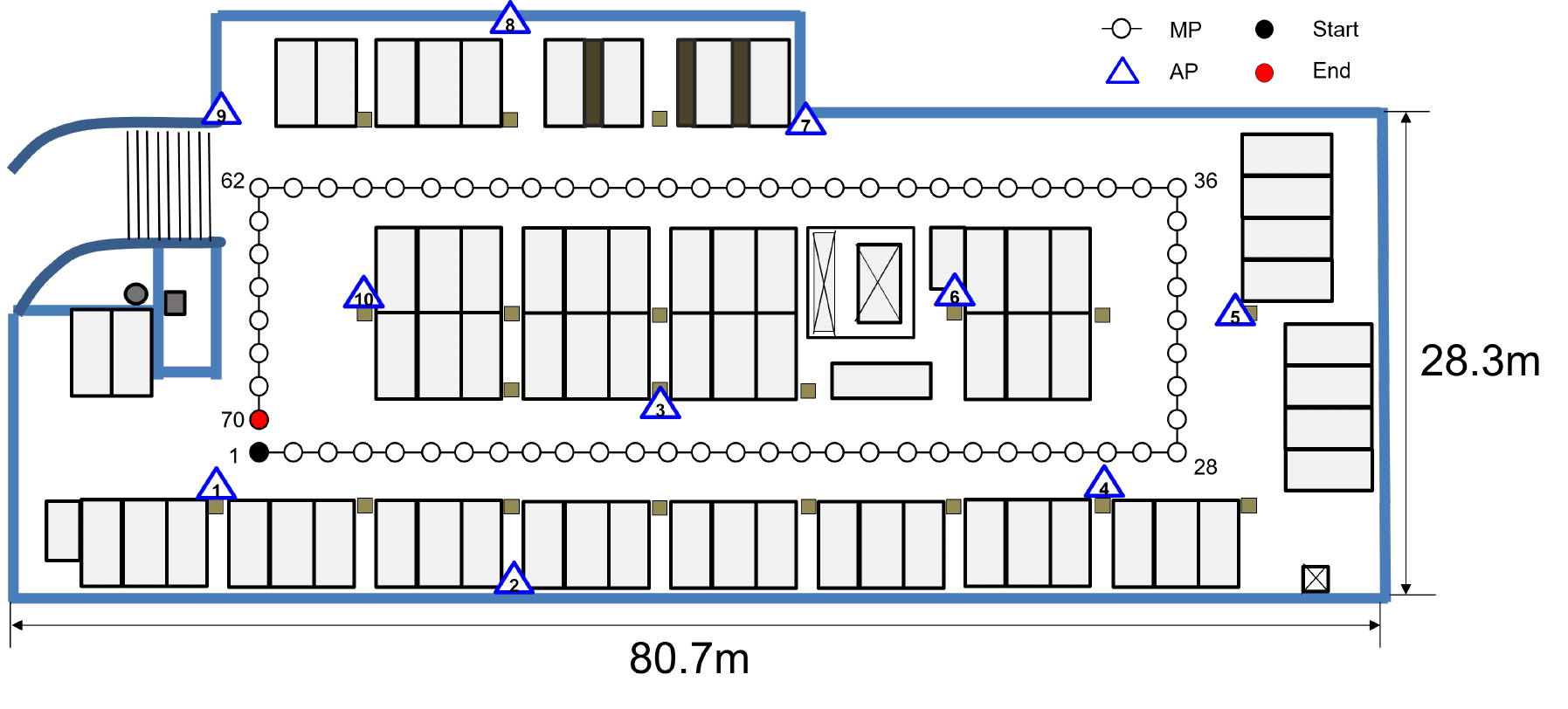}}
	\subfigure[Type $2$.]{\includegraphics[width=8.0cm]{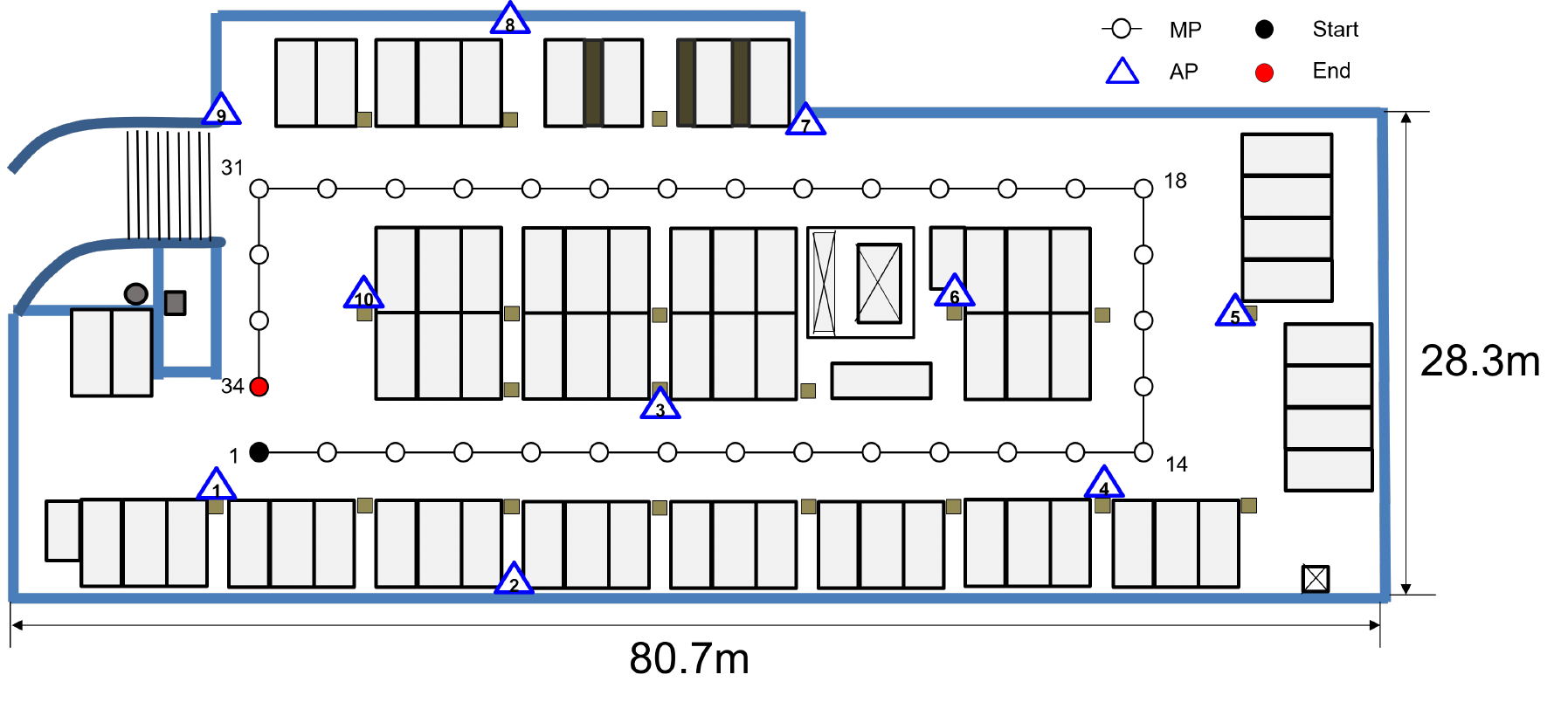}}
        \subfigure[Type $3$.]{\includegraphics[width=8.0cm]{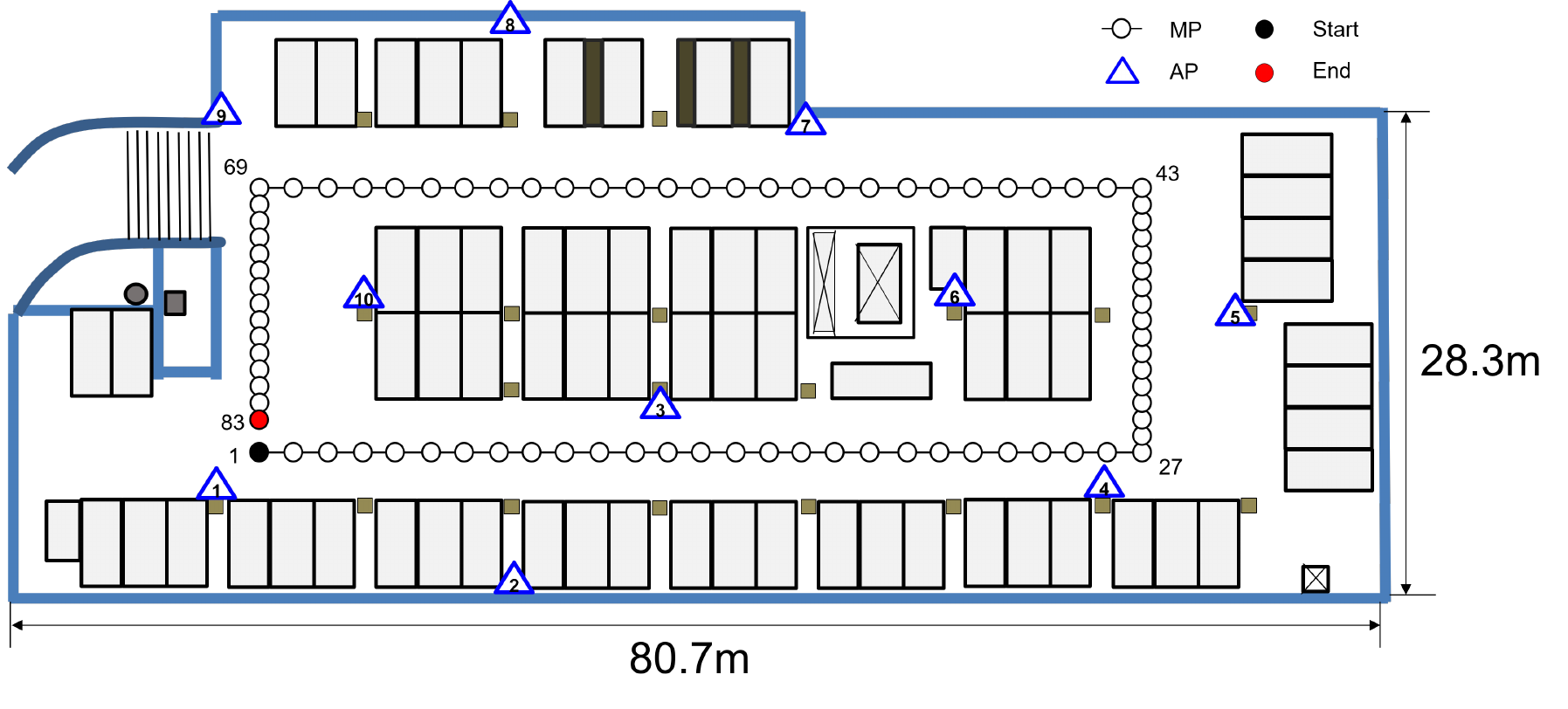}}
        \subfigure[Type $4$.]{\includegraphics[width=8.0cm]{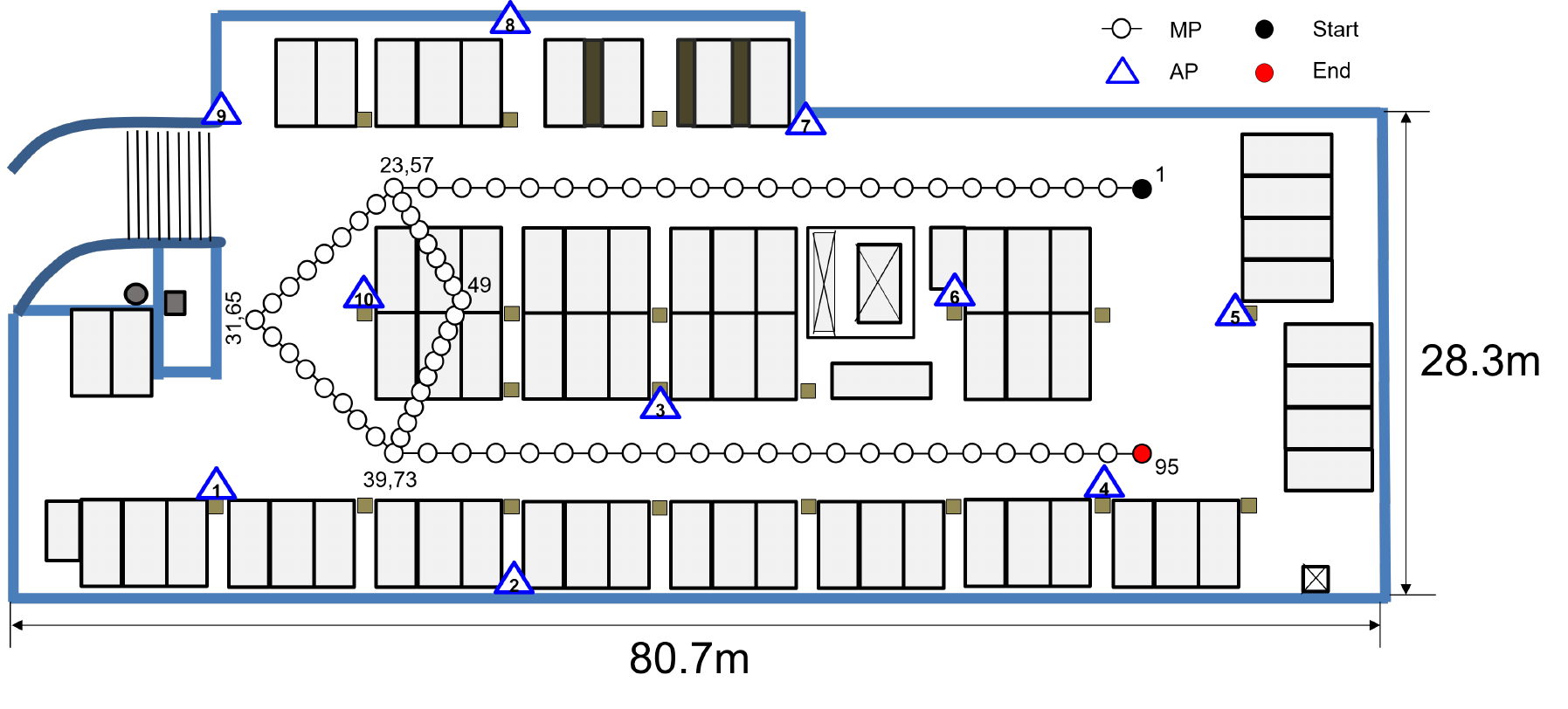}}
        \caption{Four types of user trajectories. The location of APs are represented by blue triangles, and the circles represent MPs. The detailed information of each trajectory is summarized in Table \ref{Table:Environments}. }
        \label{Fig:Path_shape1}
\end{figure*}

\begin{remark}[Simultaneous Training and Inference] \label{Remark:Train_rule}
The existing DNN-based positioning algorithms require offline training using numerous measurement data samples collected in advance. On the other hand, the proposed MINGLE is designed to train the GCN model and infer the location estimates simultaneously since it uses real-time measurements only to localize the user. This makes MINGLE attractive in practical positioning scenarios where prior measurements and their long-term statistics become invalid quickly over~time.
\end{remark}

\section{Field Experiments} \label{Sec:Verification}
This section has validated the proposed MINGLE through extensive field experiments. The experiment site and settings are explained first. Then, we compare MINGLE with several benchmarks from various perspectives. 

\subsection{Experiment Settings}\label{Subsec:Setting}

We conducted field experiments in an underground parking lot at the Korea Railroad Research Institute, Uiwang, Korea. We installed $10$ WiFi APs at the height of $2$ (\si{\metre}), which is the maximum number of WiFi APs from which one smartphone simultaneously collects RTTs. The user holds a smartphone at the height of  $1.1$ (\si{\metre}). Two kinds of smartphones (Google Pixel 2XL and Pixel 3a) are used to measure RTTs from APs.  As shown in \Cref{Fig:Test_environments}, several obstacles exist in the concerned site, such as cars and pillars, which frequently block the LoS path between the smartphone and APs.   

Four types of user mobility trajectories are considered, as represented in \Cref{Fig:Path_shape1}. Except for the last type, every trajectory follows a rectangular mobility pattern with $4$ SCs. On the other hand, the number of MPs therein are different, say $70$, $34$, and $83$ MPs for Types $1$, $2$, and $3$, respectively. Besides, the distance between adjacent MPs on Type-$1$ and Type-$2$ patterns is constant, while the Type-3 pattern has different inter-distance depending on vertical and horizontal paths to reflect different moving speeds. Contrary to the other types, the Type-$4$ trajectory comprises $8$ SCs (two SCs are overlapped) with $95$ MPs. Since only a limited time is allowed for the experiment per day, we conduct $10$ experiments for $3$ days. Table \ref{Table:Environments} gives a detailed explanation of all experiments. It is worth mentioning that the number and locations of parked cars differed for each day, resulting in a non-stationary RTT distribution, as summarized in Appendix \ref{Appendix:RTT_distribution}. In particular, RTT on Day $3$ was more corrupted than those on Days $1$ and $2$.

\begin{table}[]
    \caption{{Environments Description}} \label{Table:Environments}
    \centering
    \begin{adjustbox}{width=\textwidth/2}
\begin{tabular}{c|*{10}ccccccccccc}
Experiments \#   & 1 & 2 & 3 & 4 & 5 & 6 & 7 & 8 & 9 & 10 \\ \hline
Experiment day  & \multicolumn{1}{c|}{Day 1} & \multicolumn{3}{c|}{Day 2} & \multicolumn{6}{c}{Day 3} \\ \hline
Device         & \multicolumn{4}{c|}{Pixel 2XL} & \multicolumn{6}{c}{Pixel 3a} \\ \hline
Path shape        & \multicolumn{1}{c|}{Type 1} & \multicolumn{3}{c|}{Type 2} & \multicolumn{3}{c|}{Type 3} & \multicolumn{3}{c}{Type 4} \\ \hline

Number of node    & \multicolumn{1}{c|}{70} & 34 & 34 & \multicolumn{1}{c|}{34} & 83 & 83 & \multicolumn{1}{c|}{83} & 95 & 95 & 95 \\ \hline
Number of SCs     & \multicolumn{1}{c|}{4} & 4 & 4 & \multicolumn{1}{c|}{4} & 4 & 4 & \multicolumn{1}{c|}{4} & 8 & 8 & 8 \\
\end{tabular}
    \end{adjustbox}
\end{table}

As Remark \ref{Remark:Train_rule} mentioned, we newly trained the proposed two-layer GCN architecture for each experiment using the corresponding RTT measurement only. To avoid over-fitting, we divide all MPs' RTTs into training and validating sets with a ratio of $8:2$. The number of training epochs is up to $6000$, which can be early terminated unless the validation loss is reduced within $200$ epochs. We repeated the training process five times to obtain more reliable results by randomizing validation sets, initial weights, and seed numbers. The resultant sequence of position estimates can be derived by computing their median. We set the parameter determining the temporal adjacency $\epsilon$ specified in \eqref{Eq:Mobility_adjacency} as $2$, while the weight of the mobility regularization term $\lambda$ specified in \eqref{Eq:Loss_function} as $3$. For a semi-supervising setting, we assume that the ground-truth locations are given when the user's moving direction is changed, namely, $\alpha_n=\beta_n$ for all $n\in\mathcal{N}$.

\subsection{Performance Comparison with Benchmarks} \label{Sec:Performance_comparison}

\begin{table*} [t]
    \caption{{Summary of experiments results in Sec. \ref{Sec:Performance_comparison} (\si{\metre})}} \label{Table:Results}
    \centering
    \begin{adjustbox}{width=\textwidth}
    \begin{tabular}{cc|CCCCCCCCCC|C|CCC|C}
    \multicolumn{2}{c|}{Experiment day}             & \multicolumn{1}{c|}{Day 1} & \multicolumn{3}{c|}{Day 2} & \multicolumn{6}{c|}{Day 3}                          & \multirow{3}{*}{\makecell[c]{Total}} & \multirow{3}{*}{50th} & \multirow{3}{*}{75th} & \multirow{3}{*}{95th} & \multirow{3}{*}{{\begin{tabular}[c]{@{}c@{}}RMSE\end{tabular}}} \\ \cline{1-12}
    \multicolumn{2}{c|}{Path shape}                     & \multicolumn{1}{c|}{Type $1$} &  \multicolumn{3}{c|}{Type $2$} & \multicolumn{3}{c|}{Type $3$}                                                 & \multicolumn{3}{c|}{Type $4$} &                        &                       &                       &                       &                       \\ \cline{1-12}
    \multicolumn{1}{c|}{Learning}              & Method & 1                          & 2       & 3       & 4      & 5     & 6     & 7     & 8       & 9       & 10      &                        &                       &                       &                       &                       \\ \hline
    \multicolumn{1}{c|}{\multirow{6}{*}{{\begin{tabular}[c]{@{}c@{}}Self-\\ supervised\end{tabular}}}} & MINGLE & \mathbf{1.175}                    & \mathbf{1.373}   & \mathbf{1.445}   & \mathbf{1.211}  & \mathbf{1.894} & \mathbf{1.843} & \mathbf{1.871} & \mathbf{2.032}   & \mathbf{2.179}   & \mathbf{1.934}   & \mathbf{1.696}                  & \mathbf{1.734}                 & \mathbf{2.370}                 & \mathbf{3.319}                 & \mathbf{1.398}                 \\
    \multicolumn{1}{c|}{}                      & LLS-RS & 5.525                      & 6.250   & 6.403   & 6.771  & 7.130 & 7.777 & 7.160 & 6.034   & 6.343   & 6.587   & 6.598                  & 5.803                 & 8.861                 & 14.663                & 5.553                 \\
    \multicolumn{1}{c|}{}                      & CDA    & 1.880                      & 1.800   & 1.640   & 1.697  & 2.510 & 2.430 & 2.175 & 2.150   & 2.158   & 2.138   & 2.058                  & 1.776                 & 2.819                 & 5.366                 & 1.883                 \\
    \multicolumn{1}{c|}{}                      & TA     & 1.705                      & 1.328   & 1.724   & 2.466  & \text{-}     & \text{-}     & \text{-}     & \text{-}       & \text{-}       & \text{-}       & 1.806                  & 1.620                 & 2.515                 & 3.139                 & 1.398                 \\
    \multicolumn{1}{c|}{}                      & EKF    & 5.243                      & 3.616   & 3.716   & 3.449  & 5.077 & 5.539 & 4.923 & 4.582   & 4.708   & 5.224   & 4.608                  & 4.580                 & 6.190                 & 8.012                 & 3.639                 \\
    \multicolumn{1}{c|}{}                      & GCN    & 1.648                      & 2.993   & 3.110   & 2.974  & 2.292 & 2.515 & 2.251 & 2.460   & 2.708   & 2.532   & 2.548                  & 2.313                 & 3.225                 & 4.765                 & 1.959                 \\ \hline
    \multicolumn{1}{c|}{\multirow{2}{*}{\makecell[c]{Semi-\\supervised}}} & MINGLE & \mathbf{1.060}                      & \mathbf{1.214}   & \mathbf{1.303}   & \mathbf{1.104}  & \mathbf{1.187} & \mathbf{1.555} & \mathbf{1.477} & \mathbf{1.345}   & \mathbf{1.491}   & \mathbf{1.281}   & \mathbf{1.302}                  & \mathbf{1.224}                 & \mathbf{1.730}                 & \mathbf{2.707}                 & \mathbf{1.073}                 \\
    \multicolumn{1}{c|}{}                      & GCN    & 1.747                      & 2.991   & 3.148   & 3.001  & 2.301 & 2.620 & 2.302 & 2.602   & 2.752   & 2.656   & 2.612                  & 2.292                 & 3.290                 & 4.768                 & 2.026                
    \end{tabular}
    \end{adjustbox}
\end{table*}

\begin{figure*}[t]
    \centering 
    \subfigure[Self-supervised learning.]{\includegraphics[width=7.5cm]{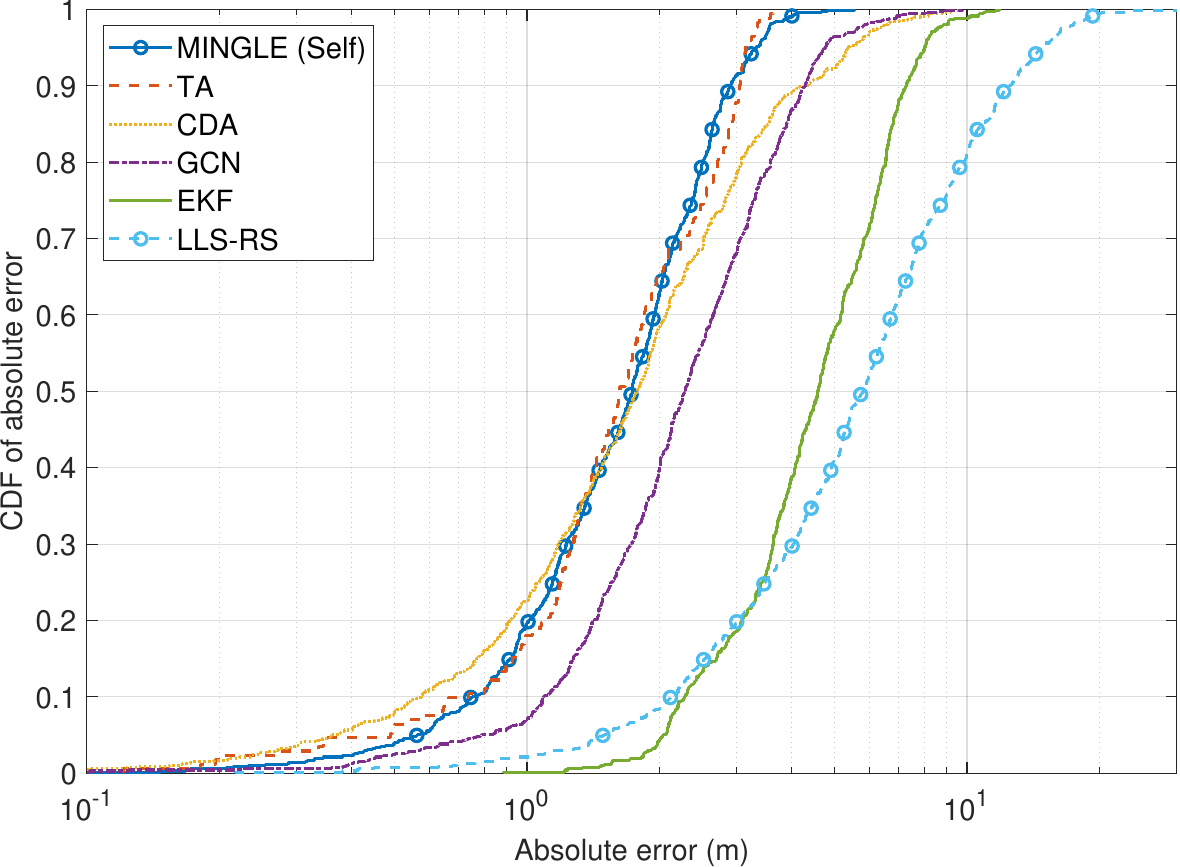} \label{Fig:CDF1}}
    \subfigure[Semi-supervised learning.]{\includegraphics[width=7.5cm]{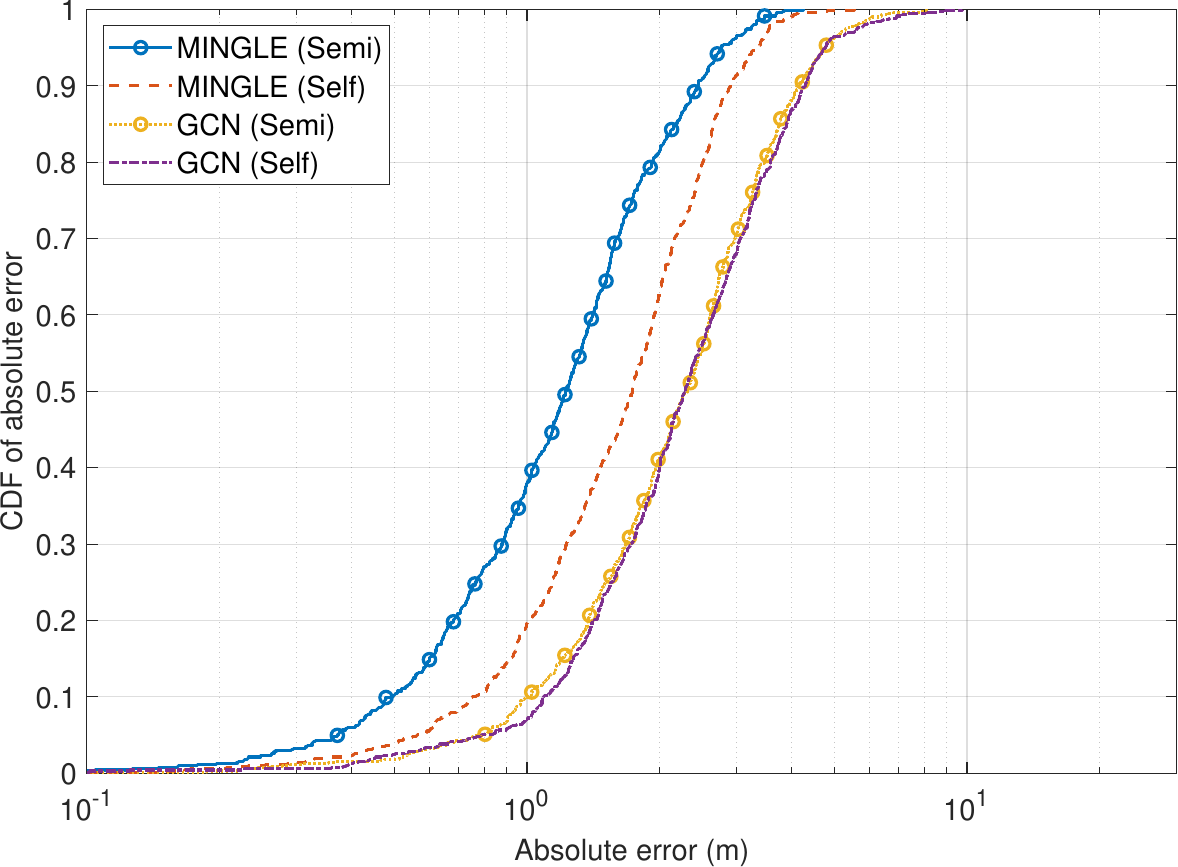} \label{Fig:CDF2} } 
    \caption{CDFs of absolute errors for MINGLE and benchmarks.}
\end{figure*}

This subsection compares MINGLE's positioning accuracy with several benchmarks introduced below. 
\begin{itemize}
 \item \emph{Linear least square through reference selection} (LLS-RS) \cite{LLS_RS2008}: One representative multilateration algorithm assuming that all RTTs are measured under LoS conditions. 
 \item CDA \cite{yu2021integrating}: Our previous work to overcome NLoS propagation by obtaining multiple PELs through different AP combinations. MINGLE uses the CDA-based location estimates as real-time labels for a self-supervised approach. The detailed process of CDA is provided in Appendix \ref{Appendix:CDA}.
 \item TA \cite{han2021exploiting}: Another previous work aligning IMU-based user trajectory into WiFi RTT measurements. Noting that TA is based on the assumption of a constant moving speed, it can be applicable only when the distance between adjacent MPs is equivalent, as in the case of Type 1 and Type 2 in \Cref{Fig:Path_shape1}.
 \item EKF \cite{sun2020indoor}: A representative fusing technique of WiFi positioning and PDR. The parameters determining the weight of WiFi and PDR results are experimentally optimized.
 \item GCN \cite{sun2021novel}: GCN-based positioning method considering all APs and MPs as nodes, and their edges are given when the corresponding inter-distance is below the threshold. The output of the GCN is the location estimates of all APs and MPs, while only a few MPs' location estimates can be involved in computing a loss function if their ground-truth labels are given.  
\end{itemize}
For a fair comparison, we compute various performance metrics, including the $50$th, $75$th, and $95$th percentiles of absolute errors, \emph{mean absolute error} (MAE), and \emph{root mean square error} (RMSE). All results are summarized in Table \ref{Table:Results}. In the following, we highlight MINGLE's superiority over benchmarks by dividing the cases into self-supervised learning (with no labels) and semi-supervised learning (with a few labels).

\begin{figure*}[t]
	\centering
	\subfigure[Type $1$.]{\includegraphics[width=4.3cm]{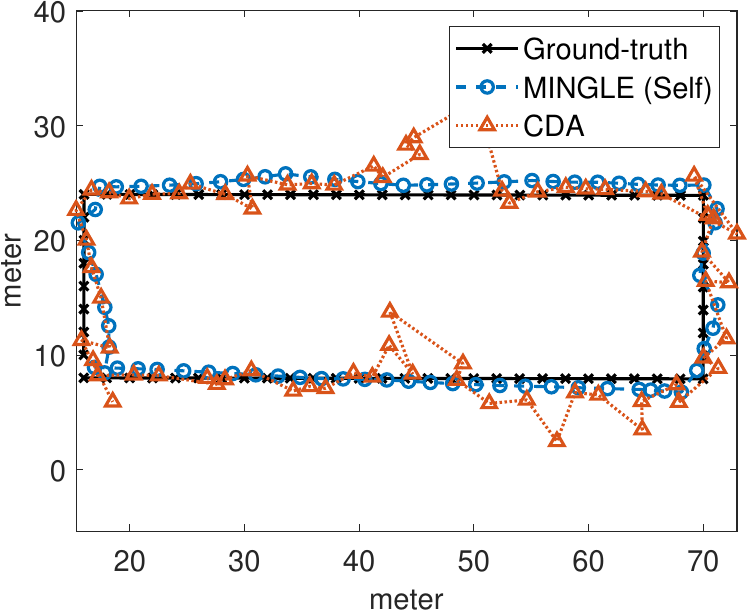}}
	\subfigure[Type $2$.]{\includegraphics[width=4.3cm]{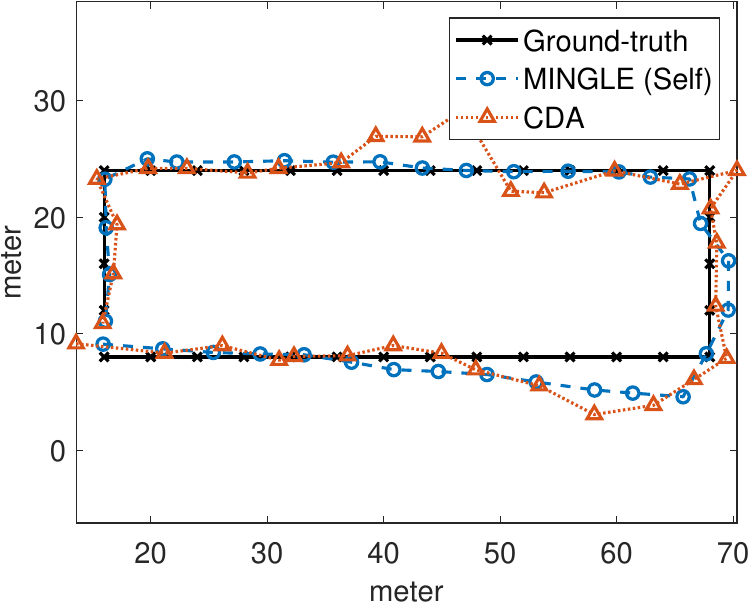}}
        \subfigure[Type $3$.]{\includegraphics[width=4.3cm]{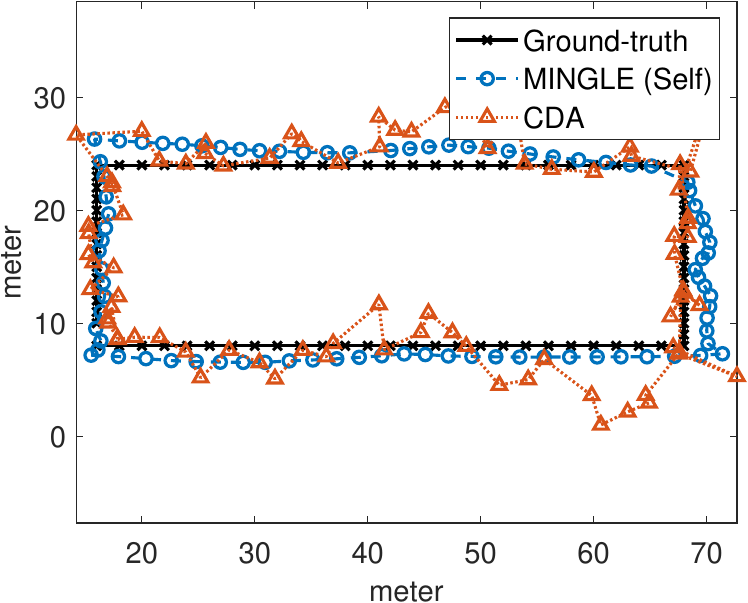}}
        \subfigure[Type $4$.]{\includegraphics[width=4.3cm]{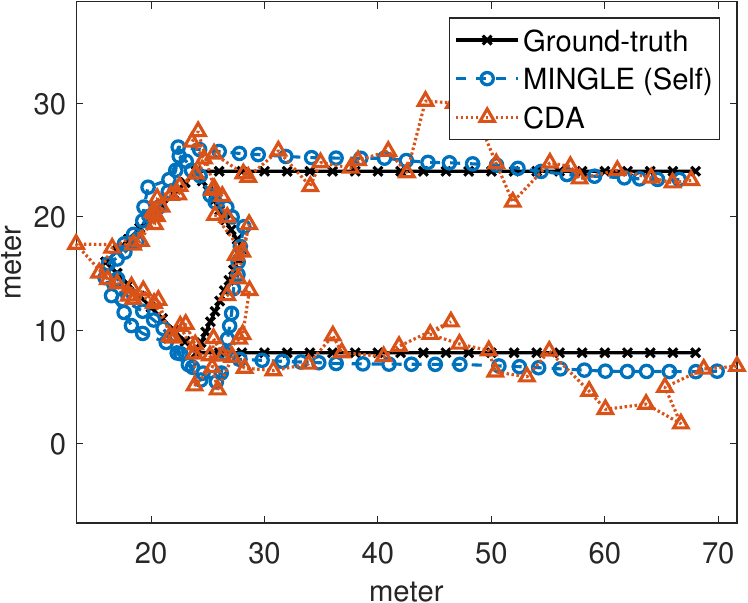}}
	\caption{Estimated user trajectories of MINGLE and CDA when no ground-truth location is given (self-supervised learning).}
	\label{Fig:MINGLE_and_CDA}
\end{figure*}

\begin{figure*}[t]
	\centering
	\subfigure[Type $1$.]{\includegraphics[width=4.3cm]{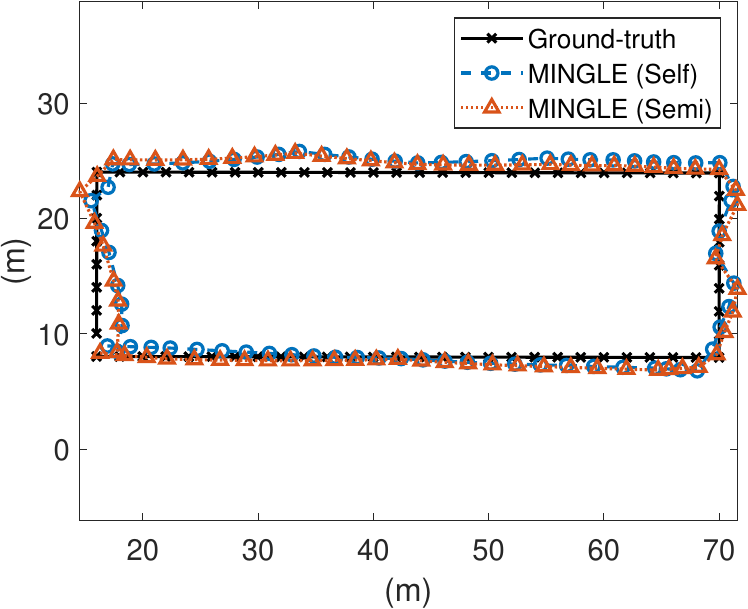}}
	\subfigure[Type $2$.]{\includegraphics[width=4.3cm]{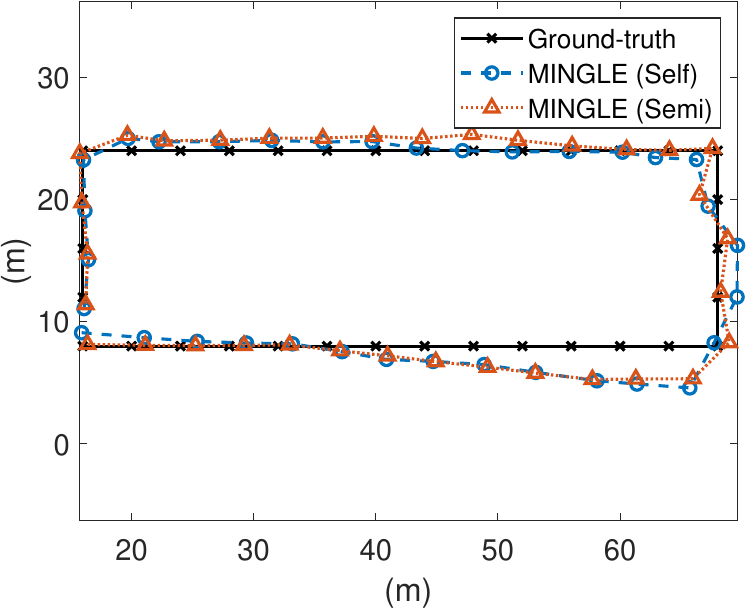}}
        \subfigure[Type $3$.]{\includegraphics[width=4.3cm]{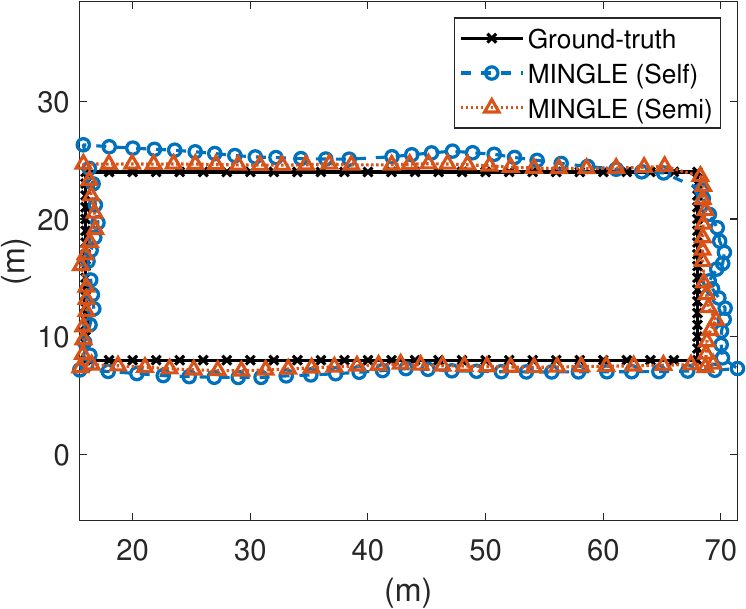}}
        \subfigure[Type $4$.]{\includegraphics[width=4.3cm]{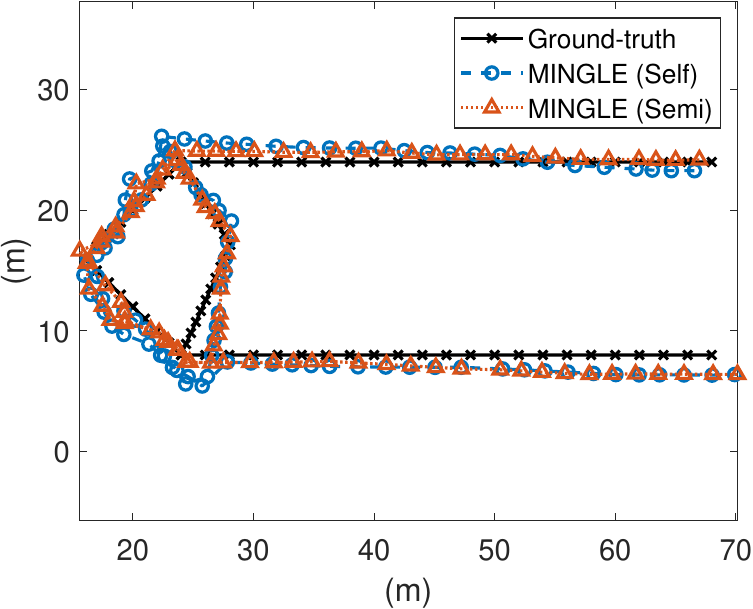}}
	\caption{Estimated user trajectories of MINGLE when a few ground-truth location is given (semi-supervised learning). We depict the result of self-supervised version of MINGLE for a fair comparison. }
	\label{Fig:Self_vs_Semi}
\end{figure*}

\subsubsection{Self-Supervised Learning} First, \Cref{Fig:CDF1} shows the \emph{cumulative distribution function} (CDF) of the absolute error when no ground-truth location is given for training. It is shown that MINGLE outperforms the others except for TA. Recalling that TA cannot be applicable for Type-$3$ and Type-$4$ trajectories due to the assumption of a constant step length, MINGLE is said to be the most effective algorithm from the perspectives of accuracy and applicability.  

One noticeable point is the comparison to CDA's results, whose location estimates are used for real-time labels of MINGLE. As observed in Fig. \ref{Fig:MINGLE_and_CDA}, 
MINGLE tends to smooth nearby location estimates by leveraging two mobility-induced graphs. On the other hand, CDA-based location estimates are derived using the corresponding RTT measurements only, prone to severe NLoS conditions. Consequently, MINGLE is effective in revising CDA's position estimates deviating from the ground truth, say the $95$th percentile of MINGLE being $3.319$ (\si{\metre}) while CDA counterpart being $5.336 $ (\si{\metre}).

\subsubsection{Semi-Supervised Learning}

\begin{figure}[t] 
    \centering 
    \includegraphics[width=6.5cm]{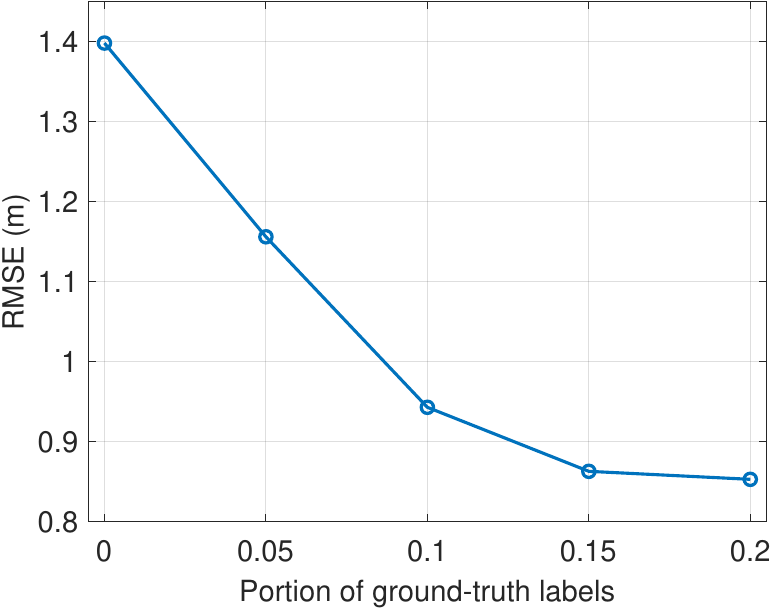} 
    \caption{RMSE vs. the portion of ground-truth labels.}
    \label{Fig:Semi_ratio}
\end{figure}

Recalling that in the case of semi-supervised learning, a few ground-truth locations are given for the MPs whose heading directions are changed, e.g., $\beta_n=1$. To verify the effect of such ground-truth labels, we compare the results of self- and semi-supervised learning in \Cref{Fig:CDF2}, showing a significant performance enhancement. As shown in \Cref{Fig:Self_vs_Semi}, MINGLE can utilize the given ground-truth locations to find the other locations, effective for Type-$3$ and Type-$4$ trajectories. We also compare the results of GCN \cite{yan2021graph} as a benchmark, revealing no performance enhancement. Eventually, MINGLE is said to be better at utilizing the ground-truth locations than the benchmark.   

Next, we aim to show the effect of number of ground-truth labels on the positioning performance. To this end, we uniformly pick $5$-$20$\% MPs on the trajectories as ground-truth labels, and measure the  RMSE of the semi-supervised version of MINGLE in \Cref{Fig:Semi_ratio}. It is observed that a small portion of labels makes the RMSE significantly reduced, say from $1.398$ (\si{\metre}) to $1.156$ (\si{\metre}) when $5$\% labels are given. On the other hand, the gain of number of the label decreases as the portion increases, and the resultant RMSE tends to converge with no more performance enhancement. 

\subsection{Cross-Graph Learning} \label{Sec:Cross-graph_validation}

As recalled in Sec. \ref{Sec:Cross_Graph}, we develop two input features, say $\mathbf{F}_1$ and $\mathbf{F}_2$, which are inputted to TMG- and DMG-based GCNs with sharing their weights.  
This subsection aims to confirm the optimality of the proposed cross-graph learning. To this end, we consider two strategies for training the network. 
\begin{itemize}
\item \textbf{Standalone Learning}: The first strategy is to train a standalone network for each individual graph. Noting that we have two inputs and two graphs, allowing us to make $4$ combinations for standalone training strategy, namely, $\mathbf{F}_1$-TMG, $\mathbf{F}_2$-TMG, $\mathbf{F}_1$-DMG, and $\mathbf{F}_2$-DMG.
\item \textbf{Cross-Graph Learning}: The second strategy is a cross-graph learning, making $4$ options, namely, ($\mathbf{F}_1$-TMG, $\mathbf{F}_1$-DMG), 
($\mathbf{F}_1$-TMG, $\mathbf{F}_2$-DMG),
($\mathbf{F}_2$-TMG, $\mathbf{F}_1$-DMG), and
($\mathbf{F}_2$-TMG, $\mathbf{F}_2$-DMG). 
\end{itemize}

All results of this study are summarized in Table \ref{Table:Best}. It is observed that our cross-graph learning strategy, say ($\mathbf{F}_1$-TMG, $\mathbf{F}_2$-DMG), provides the best positioning accuracy. The reverse assignment, say ($\mathbf{F}_2$-TMG, $\mathbf{F}_1$-DMG), also provides a similar performance with a marginal degradation. On the other hand, every standalone learning achieves a worse performance than the cross-graph counterpart, confirming the effectiveness of cross-graph learning.

\begin{table*}[t]
    \caption{{Feature-graph combination (\si{\metre}). The best combination is highlighted in bold.}} \label{Table:Best}
    \centering
    \begin{adjustbox}{width=\textwidth}
    \begin{tabular}{cccc|CCCCCCCCCC|C|CCC|C}
    \multicolumn{4}{c|}{Experiment day}                                                                                                                                         & \multicolumn{1}{c|}{\text{Day 1}} & \multicolumn{3}{c|}{\text{Day 2}}                                           & \multicolumn{6}{c|}{\text{Day 3}}                                                                          & \multirow{3}{*}{\makecell[c]{\text{Total}}} & \multirow{3}{*}{\text{50th}} & \multirow{3}{*}{\text{75th}} & \multirow{3}{*}{\text{95th}} & \multirow{3}{*}{\makecell[c]{\text{RMSE}}} \\ \cline{1-14}
    \multicolumn{4}{c|}{Path shape}                                                                                                                                               & \multicolumn{1}{c|}{Type 1} & \multicolumn{3}{c|}{Type 2} & \multicolumn{3}{c|}{Type 3}                                                                                                                         & \multicolumn{3}{c|}{Type 4}                      &                        &                       &                       &                       &                       \\ \cline{1-14}
    \multicolumn{1}{c|}{Learning}              & \multicolumn{1}{c|}{Method}                                                                  & Graph 1         & Graph 2         & 1                          & 2              & 3                                  & 4              & 5              & 6              & 7              & 8              & 9              & 10             &                        &                       &                       &                       &                       \\ \hline
    \multicolumn{1}{c|}{\multirow{8}{*}{{\begin{tabular}[c]{@{}c@{}}Self-\\ supervised\end{tabular}}}} & \multicolumn{1}{c|}{\multirow{4}{*}{\begin{tabular}[c]{@{}c@{}}Stand-\\ alone\end{tabular}}} & $\mathbf{F}_1$-TMG          & -               & 2.833                      & 2.316          & 2.273                              & 2.410          & 2.768          & 2.523          & 2.467          & 2.574          & 3.029          & 2.640          & 2.583                  & 2.475                 & 3.493                 & 5.172                 & 2.120                 \\
    \multicolumn{1}{c|}{}                      & \multicolumn{1}{c|}{}                                                                        & $\mathbf{F}_2$-TMG           & -               & 1.911                      & 1.729          & 1.728                              & 2.004          & 2.500          & 2.676          & 2.505          & 2.445          & 2.663          & 2.558          & 2.272                  & 2.128                 & 3.127                 & 4.948                 & 1.948                 \\
    \multicolumn{1}{c|}{}                      & \multicolumn{1}{c|}{}                                                                        & -               & $\mathbf{F}_1$-DMG           & 4.741                      & 5.666          & 5.101                              & 4.768          & 5.926          & 3.352          & 6.373          & 3.466          & 4.607          & 4.724          & 4.872                  & 3.717                 & 6.640                 & 12.447                & 4.287                 \\
    \multicolumn{1}{c|}{}                      & \multicolumn{1}{c|}{}                                                                        & -               & $\mathbf{F}_2$\textbf{-DMG}  & \mathbf{1.216}             & \mathbf{1.368} & \mathbf{1.455}                     & \mathbf{1.297} & \mathbf{1.886} & \mathbf{1.953} & \mathbf{1.889} & \mathbf{2.054} & \mathbf{2.334} & \mathbf{1.956} & \mathbf{1.741}         & \mathbf{1.751}        & \mathbf{2.433}        & \mathbf{3.455}        & \mathbf{1.451}        \\ \cline{2-19} 
    \multicolumn{1}{c|}{}                      & \multicolumn{1}{c|}{\multirow{4}{*}{\begin{tabular}[c]{@{}c@{}}Cross-\\ graph\end{tabular}}} & $\mathbf{F_1}$-TMG           & $\mathbf{F_1}$-DMG           & 1.591                      & 2.322          & 2.579                              & 2.300          & 2.229          & 2.548          & 2.286          & 2.404          & 2.432          & 2.378          & 2.307                  & 2.264                 & 2.944                 & 3.956                 & 1.798                 \\
    \multicolumn{1}{c|}{}                      & \multicolumn{1}{c|}{}                                                                        & $\mathbf{F_1}$\textbf{-TMG}  & $\mathbf{F_2}$\textbf{-DMG} & \mathbf{1.175}             & \mathbf{1.373} & \mathbf{1.445}                     & \mathbf{1.211} & \mathbf{1.894} & \mathbf{1.843} & \mathbf{1.871} & \mathbf{2.032} & \mathbf{2.179} & \mathbf{1.934} & \mathbf{1.696}         & \mathbf{1.734}        & \mathbf{2.370}        & \mathbf{3.319}        & \mathbf{1.398}        \\
    \multicolumn{1}{c|}{}                      & \multicolumn{1}{c|}{}                                                                        & $\mathbf{F_2}$-TMG          & $\mathbf{F_1}$-DMG           & 1.195                      & 1.336          & 1.488                              & 1.209          & 1.806          & 1.862          & 1.863          & 2.036          & 2.336          & 1.878          & 1.701                  & 1.688                 & 2.423                 & 3.358                 & 1.418                 \\
    \multicolumn{1}{c|}{}                      & \multicolumn{1}{c|}{}                                                                        & $\mathbf{F_2}$-TMG           & $\mathbf{F_2}$-DMG           & 2.844                      & 1.674          & 1.675                              & 1.822          & 2.383          & 2.577          & 2.512          & 2.396          & 2.611          & 2.475          & 2.297                  & 1.713                 & 2.791                 & 5.306                 & 1.991                 \\ \hline
    \multicolumn{1}{c|}{\multirow{8}{*}{{\begin{tabular}[c]{@{}c@{}}Semi-\\ supervised\end{tabular}}}} & \multicolumn{1}{c|}{\multirow{4}{*}{\begin{tabular}[c]{@{}c@{}}Stand-\\ alone\end{tabular}}} & $\mathbf{F_1}$-TMG           & -               & 1.698                      & 1.585          & 1.710                              & 1.604          & 2.331          & 2.567          & 2.363          & 1.947          & 2.124          & 2.052          & 1.998                  & 1.809                 & 2.836                 & 4.940                 & 1.779                 \\
    \multicolumn{1}{c|}{}                      & \multicolumn{1}{c|}{}                                                                        & $\mathbf{F_2}$-TMG           & -               & 3.123                      & 2.907          & 2.474                              & 2.131          & 2.638          & 3.168          & 2.521          & 2.070          & 1.991          & 2.587          & 2.561                  & 2.193                 & 3.498                 & 5.788                 & 2.111                 \\
    \multicolumn{1}{c|}{}                      & \multicolumn{1}{c|}{}                                                                        & -               & $\mathbf{F_1}$-DMG           & 7.229                      & 5.226          & 5.302                              & 5.026          & 4.756          & 5.430          & 5.374          & 3.829          & 4.796          & 4.598          & 5.157                  & 3.953                 & 6.982                 & 13.227                & 4.499                 \\
    \multicolumn{1}{c|}{}                      & \multicolumn{1}{c|}{}                                                                        & -               & $\mathbf{F_2}$\textbf{-DMG} & \mathbf{1.077}             & \mathbf{1.241} & \mathbf{1.639}                     & \mathbf{1.189} & \mathbf{1.612} & \mathbf{1.708} & \mathbf{1.614} & \mathbf{1.728} & \mathbf{1.728} & \mathbf{1.616} & \mathbf{1.515}         & \mathbf{1.390}        & \mathbf{2.280}        & \mathbf{3.324}        & \mathbf{1.305}        \\ \cline{2-19} 
    \multicolumn{1}{c|}{}                      & \multicolumn{1}{c|}{\multirow{4}{*}{\begin{tabular}[c]{@{}c@{}}Cross-\\ graph\end{tabular}}} & $\mathbf{F_1}$-TMG           & $\mathbf{F_1}$-DMG          & 1.756                      & 2.389          & 2.268                              & 2.146          & 2.181          & 2.192          & 2.059          & 1.871          & 1.788          & 1.978          & 2.063                  & 1.917                 & 2.743                 & 3.886                 & 1.622                 \\
    \multicolumn{1}{c|}{}                      & \multicolumn{1}{c|}{}                                                                        & $\mathbf{F_1}$\textbf{-TMG} & $\mathbf{F_2}$\textbf{-DMG}  & \mathbf{1.060}             & \mathbf{1.214} & \multicolumn{1}{l}{\textbf{1.303}} & \mathbf{1.104} & \mathbf{1.187} & \mathbf{1.555} & \mathbf{1.477} & \mathbf{1.345} & \mathbf{1.491} & \mathbf{1.281} & \mathbf{1.302}         & \mathbf{1.224}        & \mathbf{1.730}        & \mathbf{2.707}        & \mathbf{1.073}        \\
    \multicolumn{1}{c|}{}                      & \multicolumn{1}{c|}{}                                                                        & $\mathbf{F_2}$-TMG          & $\mathbf{F_1}$-DMG          & 1.065                      & 1.233          & 1.334                              & 1.111          & 1.169          & 1.505          & 1.509          & 1.335          & 1.426          & 1.257          & 1.295                  & 1.178                 & 1.744                 & 2.915                 & 1.084                 \\
    \multicolumn{1}{c|}{}                      & \multicolumn{1}{c|}{}                                                                        & $\mathbf{F_2}$-TMG           & $\mathbf{F_2}$-DMG           & 1.426                      & 1.447          & 1.340                              & 1.229          & 1.294          & 1.360          & 1.403          & 1.409          & 1.541          & 1.325          & 1.377                  & 1.265                 & 1.885                 & 2.722                 & 1.117                
    \end{tabular}
    \end{adjustbox}
\end{table*}

\begin{figure*}[t]
	\centering
	\subfigure[Self-supervised learning.]{\includegraphics[width=6.5cm]{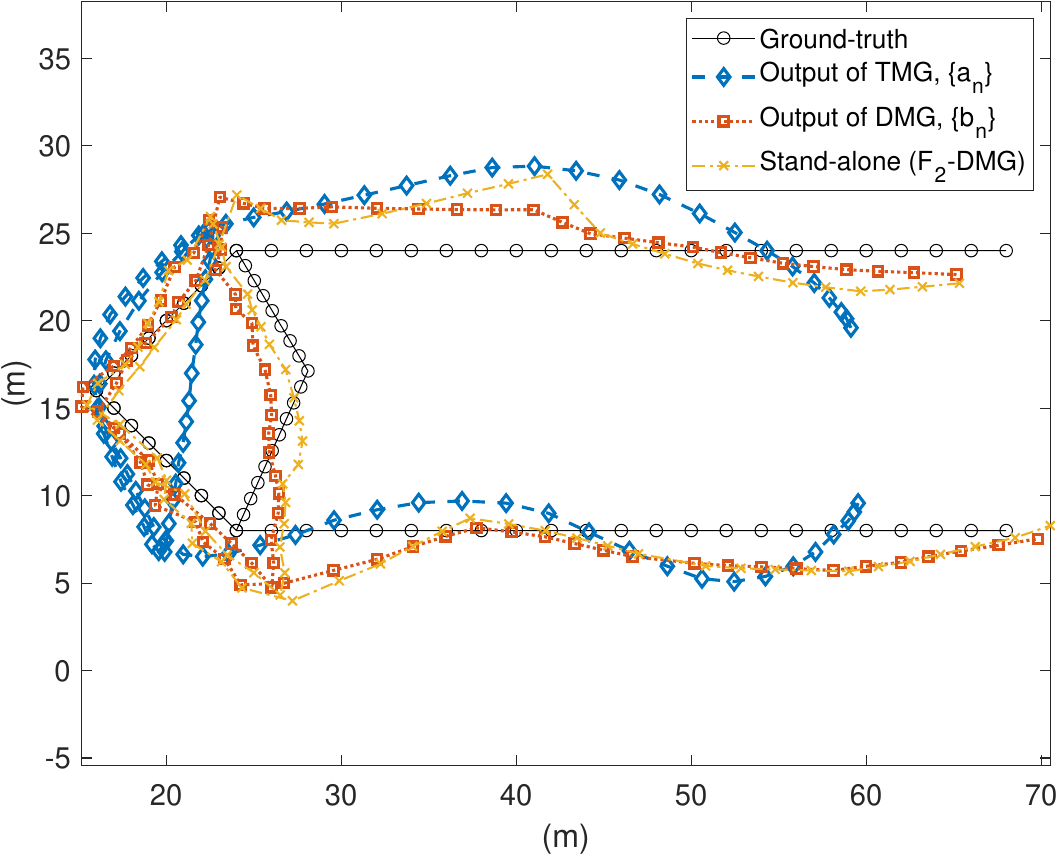}} \hspace{0.1\textwidth}
        \subfigure[Semi-supervised learning.]{\includegraphics[width=6.5cm]{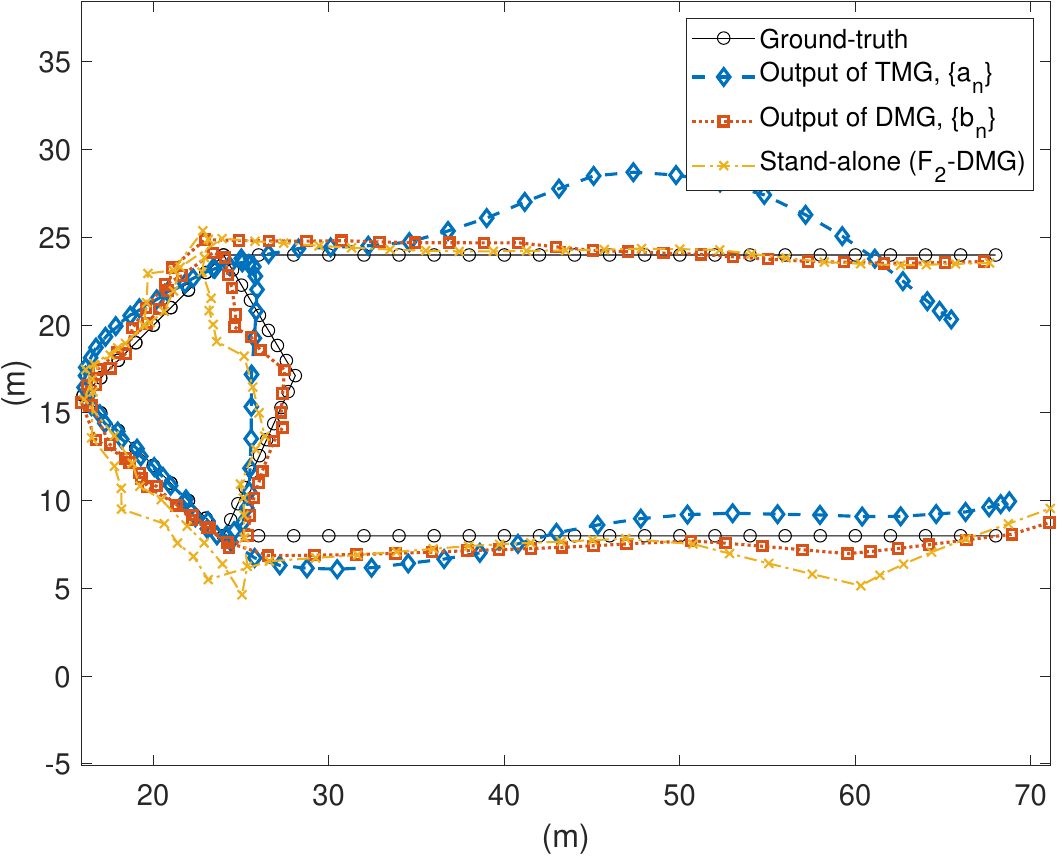}}
	\caption{Estimated user Type-4 trajectories. For cross-graph learning, the outputs of TMG and DMG, say $\{\mathbf{a}_n\}$ and $\{\mathbf{b}_n\}$ in proposed MINGLE ($\mathbf{F}_1$-TMG, $\mathbf{F}_2$-DMG), respectively, are illustrated. For standalone learning, the output of $\mathbf{F}_2$-DMG is illustrated.}
	\label{Fig:MINGLE_best_pair}
\end{figure*}

To understand each graph's contribution in detail, we illustrate the estimated Type-$4$ trajectories of our cross-graph learning in  \Cref{Fig:MINGLE_best_pair}, namely, the outputs of TMG and DMG, say $\{\mathbf{a}_n\}$ and $\{\mathbf{b}_n\}$ specified in \eqref{Eq:Vector_representations}, respectively. Besides, we add the result of the standalone learning for $\mathbf{F}_2$-DMG. It is shown that for cross-graph approach, the output of DMG plays a more crucial role in finding the location than that of TMG. It is thus reasonable to use the DMG's output as the final location estimate, as specified in \eqref{Eq:Final_estimation}. Next, from comparing the standalone approach of $\mathbf{F}_2$-DMG observing the misalignment between different SCs, the cross-graph learning approach seamlessly connects all nodes and preserves each SC's straight transition.  In summary, $\mathbf{F}_2$-DMG dominantly contributes to achieving accurate positioning, but $\mathbf{F}_1$-TMG helps improve the connectivity between SCs through cross-graph learning.

\subsection{Effect of Trajectory Length}

\begin{table*}[t]
    \caption{{MAE ($\si{\metre}$) when full-, half-, and quarter-trajectories are used. }} \label{Table:Performance_trend}
    \centering
    \begin{adjustbox}{width=\textwidth}
    \begin{tabular}{c|CCCCCCCCCC|C|CCC|C}
    Trajectory  & \text{1}     & \text{2}     & \text{3}     & \text{4}     & \text{5}     & \text{6}     & \text{7}     & \text{8}     & \text{9}     & \text{10}    & \makecell[c]{\text{Total}} & \text{50th}  & \makecell[c]{\text{75th}}  & \makecell[c]{\text{95th}}  & \makecell[c]{\text{RMSE}}  \\ \hline
    Full    & 1.175 & 1.373 & 1.445 & 1.211 & 1.894 & 1.843 & 1.871 & 2.032 & 2.179 & 1.934 & 1.696 & 1.734 & 2.370 & 3.319 & 1.398 \\
    Half   & 1.111 & 1.137 & 1.279 & 1.122 & 2.523 & 2.413 & 2.389 & 2.237 & 2.374 & 2.525 & 1.911 & 1.755 & 2.893 & 5.073 & 1.784 \\
    Quarter & 1.405 & 1.963 & 2.216 & 2.046 & 2.830 & 2.744 & 2.421 & 2.331 & 2.561 & 2.624 & 2.314 & 2.074 & 3.096 & 5.538 & 2.050
    \end{tabular}
    \end{adjustbox}
\end{table*}

To show the effect of the trajectory length on MINGLE's positioning performance, we divide each trajectory into $2$ half-trajectories and $4$ quarter-trajectories, and apply MINGLE to them. The resultant MAEs of all segmented trajectories are summarized in Table \ref{Table:Performance_trend}. It is shown that the case with an entire trajectory performs better than the other, aligned with our intuition of the number of training data samples.   
To explain, as the concerned trajectory becomes reduced, similar results among three different datasets are observed in Experiments $1$ to $4$, while a significant performance degradation is observed in Experiments $5$ to $10$. 
This decline is attributed to noisy RTT measurements at Day $3$ (See Appendix \ref{Appendix:RTT_distribution}). In other words, a longer trajectory is more beneficial to compensate for noisy measurements of WiFi RTT.

\subsection{Effect of Mobility Regularization Term} \label{Sec:Regularization}
\begin{figure*}[t]
    \centering
    \subfigure[Self-supervised learning.]{\includegraphics[width=6.5cm]{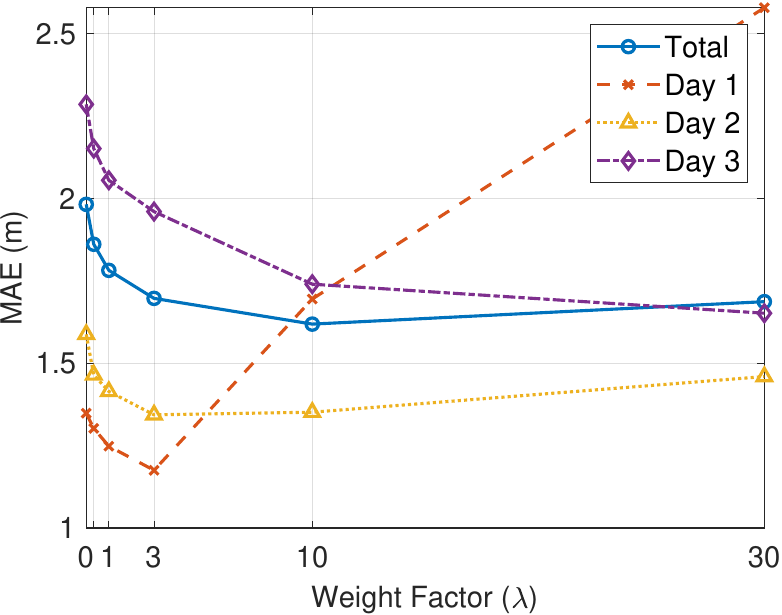}} \hspace{0.1\textwidth}
    \subfigure[Semi-supervised learning.]{\includegraphics[width=6.5cm]{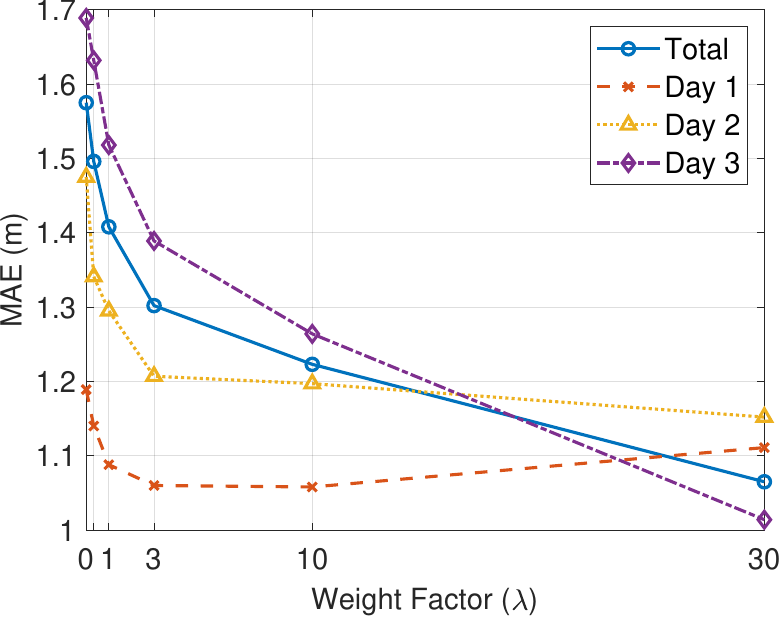}}
    \caption{Effect of mobility regularization term's weight factor $\lambda$.}
    \label{Fig:WeightFactor}
\end{figure*}

This subsection aims to check the effect of the mobility regularization term $\mathsf{Loss}_{\text{MR}}(\{\mathbf{b}_n\})$ of \eqref{Eq:Mobility_regularization} by changing its weight factor $\lambda$ from $0$ to $30$. \Cref{Fig:WeightFactor} depicts the MAE for each experiment day in terms of $\lambda$ for self- and semi-supervised learning. Noting that there is no mobility regularization when $\lambda=0$, the performance gain can be defined as the MAE gap from the minimum to the case when $\lambda=0$. Several interesting observations are made. First, in self-supervised learning, up to $0.363$ (\si{\metre}) gain is achievable through the optimization of $\lambda$ being $10$. In semi-supervised learning, on the other hand, the gain is a monotone increase of $\lambda$ in a concerned range, resulting in up to $0.510$ (\si{\metre}) gain. Second, the gain of the mobility regularization term varies depending on the experiment day; say that the enhancement of Day $3$ is the most significant, whose RTT measurements are the most noisy. Consequently, we confirm the importance of the mobility regularization term to enable self- and semi-supervised learning, and its effect becomes more dominant if a few reliable labels are added or raw RTT measurements are severely corrupted.

\section{Concluding remarks} \label{Sec:Concluding_remarks}
This paper has proposed a novel positioning technique called MINGLE, which is designed based on GNN using real-time graphs created by observing the user's mobility patterns. MINGLE has three key features to address several limitations of the existing works. First, the created mobility-induced graphs, say TMG and DMG, can be obtainable using a smartphone's low-resolution built-in IMU since they only require the IMU measurements' binary quantization values. Second, MINGLE enables us to seamlessly integrate WiFi positioning and user mobility without concern about their significance since WiFi RTT measurements and mobility information contribute to a different part of the proposed architecture. Third, CDA-based real-time labels and the mobility regularization term help design MINGLE with no or a few ground-truth labels. As a result, MINGLE is considered an effective positioning technique in practice, which is experimentally verified by showing more accurate positioning results than benchmarks.

The current work can be extended in several directions. This work considers a single-user scenario where mobility-induced graphs are created from a single user's mobility. The extension to a scenario with multiple users allows us to generate more graphs with different nodes and edges, making the problem more interesting. Next, WiFi RTT and other radio-based techniques can be integrated to make MINGLE more effective. Last, applying MINGLE to vehicle-to-everything scenarios for integrated sensing and communication will be another promising direction.

\appendix
\section{Appendix} \label{Appendix}

\subsection{RTT Error Distribution} \label{Appendix:RTT_distribution}
\begin{figure}[t]
    \centering
    \includegraphics[width=6.5cm]{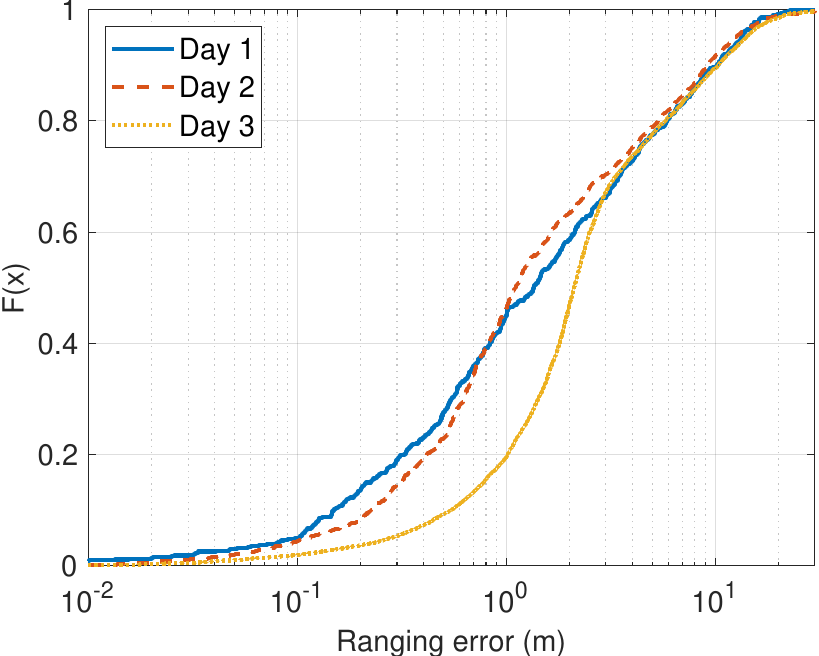}
    \caption{CDF of RTT ranging error. The number of cars parked in the experiment site and their locations cause non-stationary RTT distribution over days, resulting in different positioning performance as summarized in Table~\ref{Table:MAE_Day}.}
    \label{Fig:RTT_CDF}
\end{figure}

In our experimental settings, different scattering environments and smartphone devices can influence the RTT error distribution. We define the ranging error as the absolute error as:
\begin{align} \label{Eq:RTT_error}
    e_n^{(m)} = \left| |\mathbf{z}^{(m)}-\mathbf{x}_n| - \frac{c\cdot \tau_n^{(m)}}{2} \right|.
\end{align}
Here, $c$ represents the speed of light. We gather these errors into a set, denoted as $\{e_n^{(m)}\}$. The CDF of ranging error is visualized in Fig. \ref{Fig:RTT_CDF}. It reveals that the average ranging error on Day 3 is larger compared to those on Days 1 and 2. This leads to a degradation in performance when comparing the MAE of Days 1 and 2 to that of Day 3, as shown in Table~\ref{Table:MAE_Day}.

\begin{table}[t] 
\centering
\caption{MAE (\si{\metre}) for Experiment day} 
\begin{tabular}{c|>{\centering\arraybackslash}p{2.2cm}>{\centering\arraybackslash}p{2.2cm}}
Method        & Days 1 \& 2 & Day 3 \\ \hline
MINGLE (Self) & $1.301$          & $1.959$ \\
MINGLE (Semi) & $1.170$          & $1.389$ \\
LLS-RS        & $6.237$          & $6.839$ \\
CDA           & $1.754$          & $2.260$ \\
EKF           & $4.006$          & $5.009$
\end{tabular}
\label{Table:MAE_Day}
\end{table}

\subsection{Combinatorial Data Augmentation} \label{Appendix:CDA}
To create a self-supervised label $\mathbf{c}_n$, we utilize the CDA method described in \cite{yu2021integrating}. 
From the set of APs $\mathcal{M}$, select $K$ out of $M$ APs, and consider all possible combinations grouped as $\{\mathcal{M}_{q}\}_{q=1}^{Q}$, where $Q=\binom{M}{K}$. For a subset $\mathcal{M}_q \subset \mathcal{M}$, we define two error metrics $u_n^{(q)}$ and $v_n^{(q)}$ using RTT $\tau_n^{(m)}$ in \eqref{Eq:RTT} and PEL $\mathbf{y}^{(q)}$ in \eqref{Eq:f_n} as
\begin{align}\label{eq:error}
    u_n^{(q)} &= \sum_{m\in\mathcal{M}_q}  \left| |\mathbf{z}^{(m)}-\mathbf{y}^{(q)}| - \frac{c\cdot \tau_n^{(m)}}{2} \right|,  
    \nonumber \\ v_n^{(q)} &= \sum_{m\in \mathcal{M}_q} \tau_n^{(m)}. 
\end{align}
Next, we define a set of PELs as $\mathcal{Z}_n=\{\mathbf{y}^{(q)}\}_{q=1}^{Q}\in \mathbb{R}^{Q \times 2}$. Based on this set of PELs, the author of \cite{yu2021integrating} proposed two filtered sets, denoted by $\mathcal{Z}_n^{(1)}$ and $\mathcal{Z}_n^{(2)}$, using the above error metrics as
\begin{align}
    \mathcal{Z}_n^{(1)} &= \{ \mathbf{y}^{(q)} \in \mathcal{Z}_n \mid u_n^{(q)} \leq u_n^{(q_1)} \}, \nonumber \\
    \mathcal{Z}_n^{(2)} &= \{ \mathbf{y}^{(q)} \in \mathcal{Z}_n^{(1)} \mid v_n^{(q)} \leq v_n^{(q_2)} \},
\end{align}
where $q_1$ and $q_2$ represent the indices of PELs with the $Q_1$-th and $Q_2$-th lowest error among $\mathcal{Z}_n$ and $\mathcal{Z}_n^{(1)}$, respectively. Finally, the estimated position $\mathbf{c}_n$ is calculated as the median of $\mathcal{Z}_n^{(2)}$:
\begin{align}
    \mathbf{c}_n = \text{Median}(\mathcal{Z}_n^{(2)}).
\end{align}
In our experiments, we set $K=3$, $Q_1=37$, and $Q_2=12$ for Experiments $1$ to $4$, and $K=3$, $Q_1=40$, and $Q_2=20$ for Experiments $5$ to $10$, respectively. These parameters were optimized using numerical methods.

\balance

\def\bibfont{\footnotesize}
\bibliographystyle{IEEEtran}
\bibliography{references}

\end{document}